\documentclass{article}
\usepackage{amssymb}

\usepackage[final]{corl_2025} 

\usepackage{multirow}
\usepackage{amsmath}
\usepackage{siunitx}
\usepackage{graphicx}
\usepackage{booktabs}
\usepackage{array}
\usepackage{xcolor}
\usepackage{caption}
\usepackage{subcaption}  

\usepackage{algorithm}
\usepackage{algorithmic}
\usepackage{enumitem}
\usepackage{adjustbox} 
\usepackage{hyperref}

\hypersetup{
    colorlinks=true,
    urlcolor=magenta,
}

\title{CARE: Enhancing Safety of Visual Navigation through Collision Avoidance via Repulsive Estimation}

\author{
\textbf{Joonkyung Kim}$^{\dagger 1}$, 
\textbf{Joonyeol Sim}$^{\dagger 1}$, 
Woojun Kim$^{2}$, 
Katia Sycara$^{2}$, 
Changjoo Nam$^{1}$ \\
$^1$Department of Electronic Engineering, Sogang University\\
$^2$Robotics Institute, Carnegie Mellon University \\
$^{\dagger}$\small Equal contribution
}

\begin{document}
\maketitle

\vspace{-1em}
\begin{abstract}
We propose CARE (Collision Avoidance via Repulsive Estimation) to improve the robustness of learning-based visual navigation methods. Recently, visual navigation models, particularly foundation models, have demonstrated promising performance by generating viable trajectories using only RGB images. However, these policies can generalize poorly to environments containing out-of-distribution (OOD) scenes characterized by unseen objects or different camera setups (e.g., variations in field of view, camera pose, or focal length). Without fine-tuning, such models could produce trajectories that lead to collisions, necessitating substantial efforts in data collection and additional training.
To address this limitation, we introduce CARE, an attachable module that enhances the safety of visual navigation without requiring additional range sensors or fine-tuning of pretrained models. CARE can be integrated seamlessly into any RGB-based navigation model that generates local robot trajectories. It dynamically adjusts trajectories produced by a pretrained model using repulsive force vectors computed from depth images estimated directly from RGB inputs. 
We evaluate CARE by integrating it with state-of-the-art visual navigation models across diverse robot platforms. Real-world experiments show that CARE significantly reduces collisions (up to 100\%) without compromising navigation performance in goal-conditioned navigation, and further improves collision-free travel distance (up to 10.7×) in exploration tasks. Project page: \href{https://airlab-sogang.github.io/CARE/}{\textcolor{magenta}{https://airlab-sogang.github.io/CARE/}}




\end{abstract}

\keywords{Vision-based Navigation, Reactive Planning, Collision Avoidance}

\vspace{-1em}
\begin{figure}[ht]
    \centering
    \includegraphics[width=1.0 \linewidth]{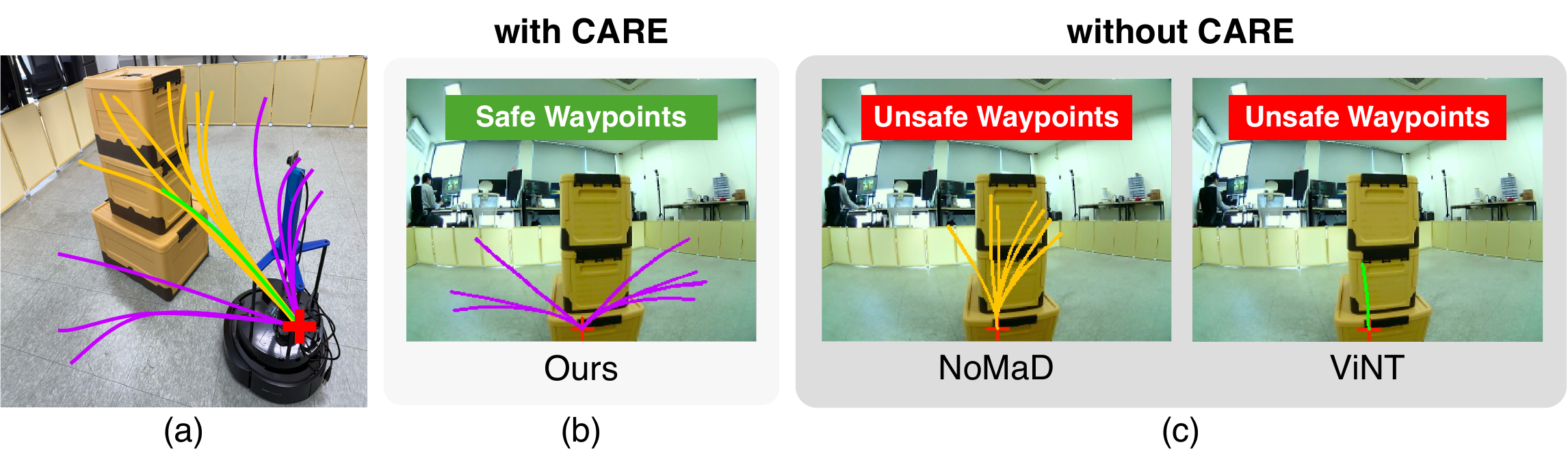}
    \vspace{-1em}
    \caption{Comparison of trajectory outputs under out-of-distribution (OOD) obstacle settings. \textbf{(a)} A test environment with trajectories with and without CARE. \textbf{(b)} Trajectories adjusted using CARE to avoid collision. \textbf{(c)} Trajectories from the original models without CARE result in collisions.}
    \label{fig:compare}
\end{figure}
\vspace{-1em}

\section{Introduction}
\vspace{-1em}
Recent advances in image representation learning and large language models (LLMs) have led to the development of vision-language models (VLMs) capable of contextual reasoning and high-level instruction following~\cite{vision_survey}. These models enable robust robotic navigation across varied tasks with minimal task-specific engineering, achieving strong performance even in few-shot~\cite{llm-planner} or zero-shot~\cite{navgpt} settings. VLMs further allow robots to localize goals from language~\cite{vlmap} and perform long-horizon navigation~\cite{lm-nav} based on human-language instructions.

To extend such high-level generalization to low-level control on physical robots, vision-based navigation models have been developed to handle variations in camera intrinsics, robot dynamics, and hardware configurations~\cite{ving}. Foundation models such as GNM~\cite{gnm} and ViNT~\cite{vint} are designed for this purpose, enabling generalization across heterogeneous platforms with minimal fine-tuning. These models support a range of tasks including goal-directed navigation, long-horizon planning with high-level policies~\cite{viking}, and exploration~\cite{nomad}, relying solely on monocular RGB input.


Despite their generalization capabilities, vision-based foundation models face safety limitations, particularly in out-of-distribution (OOD) environments (see Figure~\ref{fig:compare}). As they rely on appearance-based reasoning without explicit geometric understanding, they may produce unsafe trajectories that lead to collisions~\cite{mononav}. Adapting them to new robot platforms or camera setups often requires retraining and data collection~\cite{foundation}, limiting their real-world applicability.

To address these limitations, we propose CARE (Collision Avoidance via Repulsive Estimation), a plug-and-play safety module that enhances vision-based navigation without additional sensors or fine-tuning. CARE combines monocular depth estimation with reactive local planning to adjust trajectories from pretrained policies. It uses a pretrained monocular depth model (in this work, UniDepthV2~\cite{unidepthv2}) to infer a top-down obstacle map from RGB input, and applies Artificial Potential Fields (APF)~\cite{apf} to compute repulsive forces that steer trajectories away from obstacles. CARE can be integrated with any RGB-based navigation policy, improving safety while minimally deviating from the original path. We validate CARE on three distinct robot platforms (LoCoBot, TurtleBot4, and RoboMaster), demonstrating consistent collision reduction in unseen environments and improved collision-free distance, without compromising navigation performance.

Our contributions are summarized as follows.  
\textbf{(i)} We introduce CARE, a plug-and-play module that improves collision avoidance by integrating vision-based navigation model with reactive planning based on monocular RGB input, without requiring additional sensors or retraining.  
\textbf{(ii)} CARE generalizes across diverse vision-based navigation models, robot platforms, and camera configurations, while preserving navigation performance.  
\textbf{(iii)} We extensively validate CARE in real-world experiments across three robot platforms, achieving up to 100\% collision reduction in goal-conditioned navigation and up to 10.7× improvement in collision-free travel distance during exploration.
\vspace{-6pt}

\section{Related Work}
\label{sec:related_work}
\vspace{-1em}
Recent advances in vision-language and vision-based navigation have enabled robots to perform complex tasks using only RGB inputs. However, ensuring safe operation in dynamic and unfamiliar environments remains a key challenge. We review prior work on vision-based navigation and recent efforts to improve safety in learning-based navigation.


VLM-based navigation approaches such as LM-Nav~\cite{lm-nav}, LFG~\cite{lfg}, and NavGPT~\cite{navgpt} leverage pretrained VLMs to interpret natural language commands and perform high-level planning without task-specific supervision. VLMaps~\cite{vlmap} integrates vision-language features into 3D spatial maps, while Obstructed VLN~\cite{obvln} and Knowledge-Enhanced Scene Understanding~\cite{kesu} aim to improve robustness to domain shifts. 
However, these methods mainly focus on high-level semantic reasoning and rely on discrete or graph-based action spaces, leaving safety-critical aspects such as collision avoidance largely unaddressed.


To bridge the gap between abstract high-level commands and low-level control, recent efforts have developed vision-based foundation models for navigation that directly map RGB observations to robot actions. ViKiNG~\cite{viking} combined learned visual traversability with geographic hints for kilometer-scale navigation, while ViNG~\cite{ving} introduced topological navigation without requiring spatial maps. GNM~\cite{gnm} and ViNT~\cite{vint} further advanced by enabling generalization across heterogeneous robot platforms through minimal fine-tuning, leveraging embodiment-agnostic policies or Transformer~\cite{transformer} architectures. NoMaD~\cite{nomad} explored diffusion-based multimodal action generation, and Navigation World Models (NWM)~\cite{nwm} proposed leveraging video prediction for policy planning without explicit mapping. Despite impressive generalization, these models primarily rely on vision-only reasoning without explicit geometric understanding. This makes it difficult to guarantee safety or reactive obstacle avoidance, especially when encountering unseen obstacles or when precise spatial reasoning is required. Moreover, although they are designed for minimal fine-tuning, adapting them to new robot platforms, camera intrinsics, or dynamic environments often demands nontrivial data collection and retraining~\cite{foundation}, limiting their practical applicability.



Ensuring reliable collision avoidance has therefore emerged as a key concern in vision-based navigation. Safe-VLN~\cite{safevln} introduces waypoint filtering and fallback strategies using simulated 2D-LiDAR. Failure Prediction~\cite{failure_pred} estimates risk from visual input, while PwC~\cite{pwc} calibrates perception models for robustness under domain shift. Adaptive methods such as online constraint updates~\cite{languageupdate}, control barrier functions~\cite{asma}, and velocity-obstacle-based shielding~\cite{navrl} have also been explored. MonoNav~\cite{mononav} reconstructs local 3D metric maps from monocular depth to enable safer planning. However, its conservative obstacle avoidance strategy often sacrifices task success, leading to lower completion rates compared to other vision-based models. Moreover, it assumes static environments, limiting performance in dynamic scenarios. Similarly, NavRL~\cite{navrl} integrates a velocity-obstacle-based safety shield on top of reinforcement learning policies, but still depends on retraining and policy adaptation for different robot platforms.

In contrast, we propose CARE, an attachable module that integrates the outputs of vision-based navigation models and monocular depth estimation with a reactive planning method based on APF, a widely used approach for collision avoidance using local observations with low computational cost~\cite{apf-switch, apf_rl, rpf, dacoop}. CARE dynamically adjusts planned trajectories using the repulsive force formulation of APF, enabling safer, reactive trajectory adaptation across diverse robotic platforms without additional sensors, explicit 3D reconstruction, or fine-tuning of the navigation model.
\vspace{-1em}


\begin{figure}[t!]
    \centering
    \includegraphics[width=1.0\linewidth]{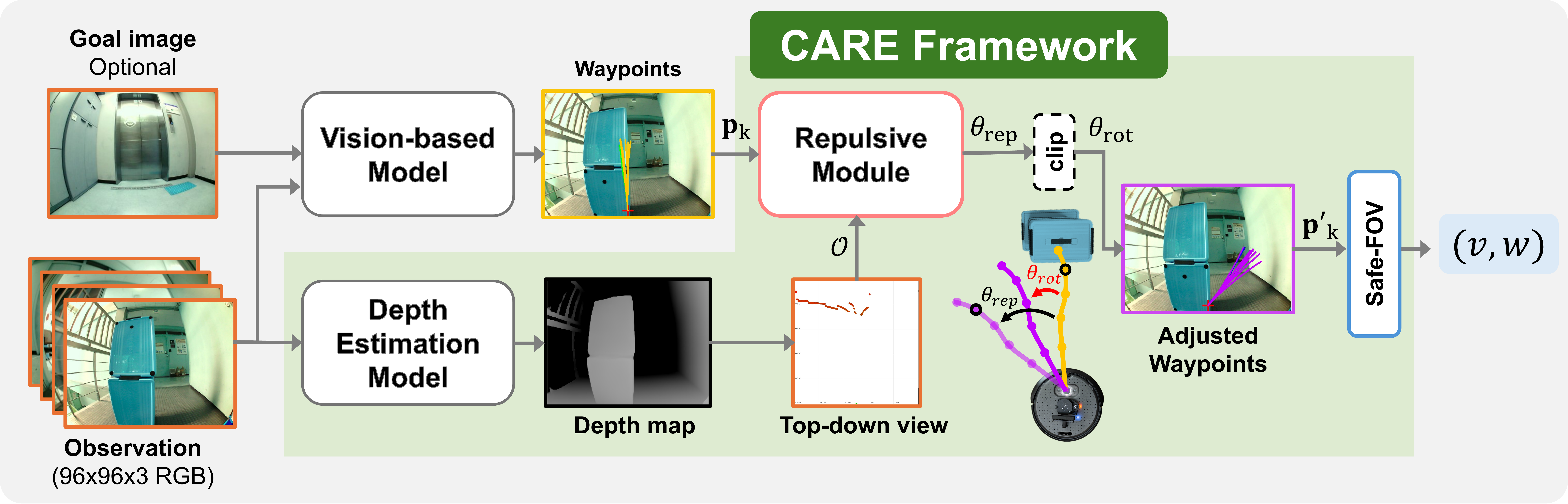}
    \caption{Overview of CARE. The system takes RGB observations and optionally a goal image as input. A vision-based model (e.g., NoMaD~\cite{nomad}, ViNT~\cite{vint}) generates waypoints or trajectories, while a top-down local map is generated from depth map. A repulsive module then adjusts the trajectory to avoid obstacles, with the adjustment angle $\theta_\text{rep}$ clipped to $\theta_\text{rot}$ for stable navigation.}
    \label{fig:Overview}
    \vspace{-0.4em}
\end{figure}

\section{Collision Avoidance via Repulsive Estimation (CARE)}
\label{sec:method}
\vspace{-0.9em}

\begin{algorithm}[t!]
\caption{CARE: Collision Avoidance via Repulsive Estimation}
\label{alg:care}
\begin{algorithmic}[1]
\renewcommand{\algorithmicrequire}{\textbf{Input:} } 
\renewcommand{\algorithmicensure}{\textbf{Output: }}
\REQUIRE RGB observation $I$, trajectory $\mathcal{T}$, depth estimation model $\mathcal{D}$
\ENSURE Control command $(v, \omega)$
\STATE $\mathcal{O} \gets \textit{ConstructTopDownObstacleMap}(I, \mathcal{D})$ \hfill \textcolor{gray}{// Stage 1: Top-down range estimation}\label{alg:construct}
\STATE $\theta_{\text{rep}} \gets \textit{EstimateRepulsiveDirection}(\mathcal{T}, \mathcal{O})$ \hfill \textcolor{gray}{// Stage 2: Repulsive force estimation}\label{alg:estimate}
\STATE $\mathcal{T}' \gets \textit{RotateTrajectory}(\mathcal{T}, \theta_{\text{rep}})$\label{alg:rotate}
\STATE $\theta_{\text{des}} \gets \textit{ComputeDesiredHeading}(\mathcal{T}')$ \hfill \textcolor{gray}{// Stage 3: Safety-enhancing mechanism}\label{alg:compute}
\IF{$|\theta_{\text{des}}| > \theta_{\text{thres}}$}\label{alg:safetybegin}
    \STATE $(v, \omega) \gets \textit{RotateInPlace}(\theta_{\text{des}})$\label{alg:rotateinplace}
\ELSE
    \STATE $(v, \omega) \gets \textit{MoveForwardAndTurn}(\theta_{\text{des}})$\label{alg:moveforward}
\ENDIF\label{alg:safetyend}
\STATE \textbf{return} $(v, \omega)$
\end{algorithmic}
\end{algorithm}


CARE is a plug-and-play module that augments the output of any vision-based navigation policy producing waypoints or trajectories. It uses the same RGB observation as the navigation model to estimate depth and adjust the predicted path accordingly. CARE operates in three stages (see Figure~\ref{fig:Overview} and Algorithm~\ref{alg:care}), enabling real-time correction of unsafe paths when encountering previously unseen obstacles.
\vspace{-5pt}


\subsection{Top-Down Range Estimation}
\label{sec:prelim_topdown}
\vspace{-3pt}

CARE first constructs a top-down local map (see Figure~\ref{fig:top-down}) by predicting the scene geometry from a depth image predicted from an RGB observation through the function \textit{ConstructTopDownObstacleMap}$(I, \mathcal{D})$ (see Algorithm~\ref{alg:care}, Line~\ref{alg:construct}).

Given an RGB observation $I \in \mathbb{R}^{H \times W \times 3}$ captured by a calibrated monocular camera, a pretrained metric depth model predicts a dense depth map. This depth map can be projected into a set of three-dimensional points $(X_i, Y_i, Z_i) \in \mathbb{R}^3$ by applying the inverse intrinsic matrix to each pixel location~\cite{peudo-lidar, unidepth, monopp}. In our implementation, we employ UniDepthV2~\cite{unidepthv2}, which directly outputs 3D point clouds without requiring explicit back-projection, though CARE can be replaced with any absolute depth estimation model capable of producing similar output formats.

The predicted point cloud $\mathcal{P} = \{ (X_i, Y_i, Z_i) \mid i = 1, \ldots, N \}$ is filtered to retain only points satisfying $(Z > 0) \land (Z \leq \tau_z) \land (Y \geq -\epsilon)$, where $\tau_z \in \mathbb{R}$ denotes the maximum sensing range and $\epsilon \in \mathbb{R}$ defines a vertical margin to exclude ceiling points. The resulting set of valid projected points is defined as $\mathcal{P}_{\text{valid}} = \{ (X_i, Z_i) \mid (X_i, Y_i, Z_i) \in \mathcal{P},\; \text{Mask}(X_i, Y_i, Z_i) = \text{True} \}$.

The $x$-axis is discretized into $M$ uniform bins, and for each bin, the closest point along the $Z$-axis is selected:
\[
(x^\star, z^\star) = \underset{(x,z) \in \mathcal{P}_{\text{valid}}}{\mathrm{argmin}} \, z,
\]
forming the obstacle set $\mathcal{O} = \{ (x^\star_j, z^\star_j) \mid j = 1, \ldots, M \}$. All coordinates are represented in the local robot frame where the positive $x$-axis points forward and the positive $y$-axis points to the left. \vspace{-3pt}

\subsection{Repulsive Force Estimation and Trajectory Adjustment}
\label{sec:repulsive}
\vspace{-3pt}



To adjust trajectories in response to nearby obstacles, CARE estimates repulsive forces using the APF method. This repulsive vector provides a reactive adjustment direction to steer the robot away from collisions.

Given the obstacle set $\mathcal{O}$ derived from monocular depth estimation, CARE computes repulsive forces applied to a generated trajectory $\mathcal{T} = \{ \mathbf{p}_1, \ldots, \mathbf{p}_K \}$ or a single waypoint $\mathbf{p}_k \in \mathbb{R}^2$, depending on the output format of the vision policy. We describe the method assuming a trajectory-based policy such as ViNT or NoMaD, which outputs a sequence of $K$ waypoints $\mathcal{T} = \{ \mathbf{p}_1, \ldots, \mathbf{p}_K \}$ at each step, though the procedure naturally extends to single-waypoint outputs.

The repulsive force at each waypoint is computed via \textit{EstimateRepulsiveDirection}$(\mathcal{T}, \mathcal{O})$ (Algorithm~\ref{alg:care}, Line~\ref{alg:estimate}), following the same formulation used in a recent work on APF with limited sensing range~\cite{apf-switch}: 
\begin{equation}
\mathbf{F}_{\text{rep}}(\mathbf{p}_k, \mathcal{O}) = \sum_{m=1}^M \frac{-1}{\|\mathbf{p}_k - \mathbf{o}_m\|_2^3} \cdot \frac{\mathbf{p}_k - \mathbf{o}_m}{\|\mathbf{p}_k - \mathbf{o}_m\|_2},
\label{eq:repulsive}
\end{equation}
where $\mathbf{o}_m \in \mathbb{R}^2$ denotes the $m$-th obstacle in $\mathcal{O}$ and $\|\cdot\|_2$ is the Euclidean norm.

To determine the adjustment amount and direction, CARE selects the waypoint experiencing the maximum repulsive magnitude:
\begin{equation}
k^\star = \mathrm{argmax}_{k \in \{1, \ldots, K\}} \| \mathbf{F}_{\text{rep}}(\mathbf{p}_k, \mathcal{O}) \|_2.
\label{eq:maxrep}
\end{equation}
For waypoint-only policies, the single predicted point is directly used in place of $\mathbf{p}_{k^\star}$. The repulsive vector at $\mathbf{p}_{k^\star}$ is converted to a heading adjustment angle $\theta_{\text{rep}} = \arctan2(F_{\text{rep},y}(\mathbf{p}_{k^\star}), F_{\text{rep},x}(\mathbf{p}_{k^\star}))$, which determines the direction of deviation from the current trajectory. 
To avoid excessive deviation from the goal, the adjustment angle is clipped as $\theta_{\text{rot}} = \text{clip}(\theta_{\text{rep}}, -\theta_{\text{clip}}, \theta_{\text{clip}})$, where $\text{clip}(\cdot, a, b)$ limits the value between $a$ and $b$.

\begin{figure}[t!]
    \centering
    \begin{subfigure}[t]{0.45\linewidth}
        \centering
        \includegraphics[width=\linewidth]{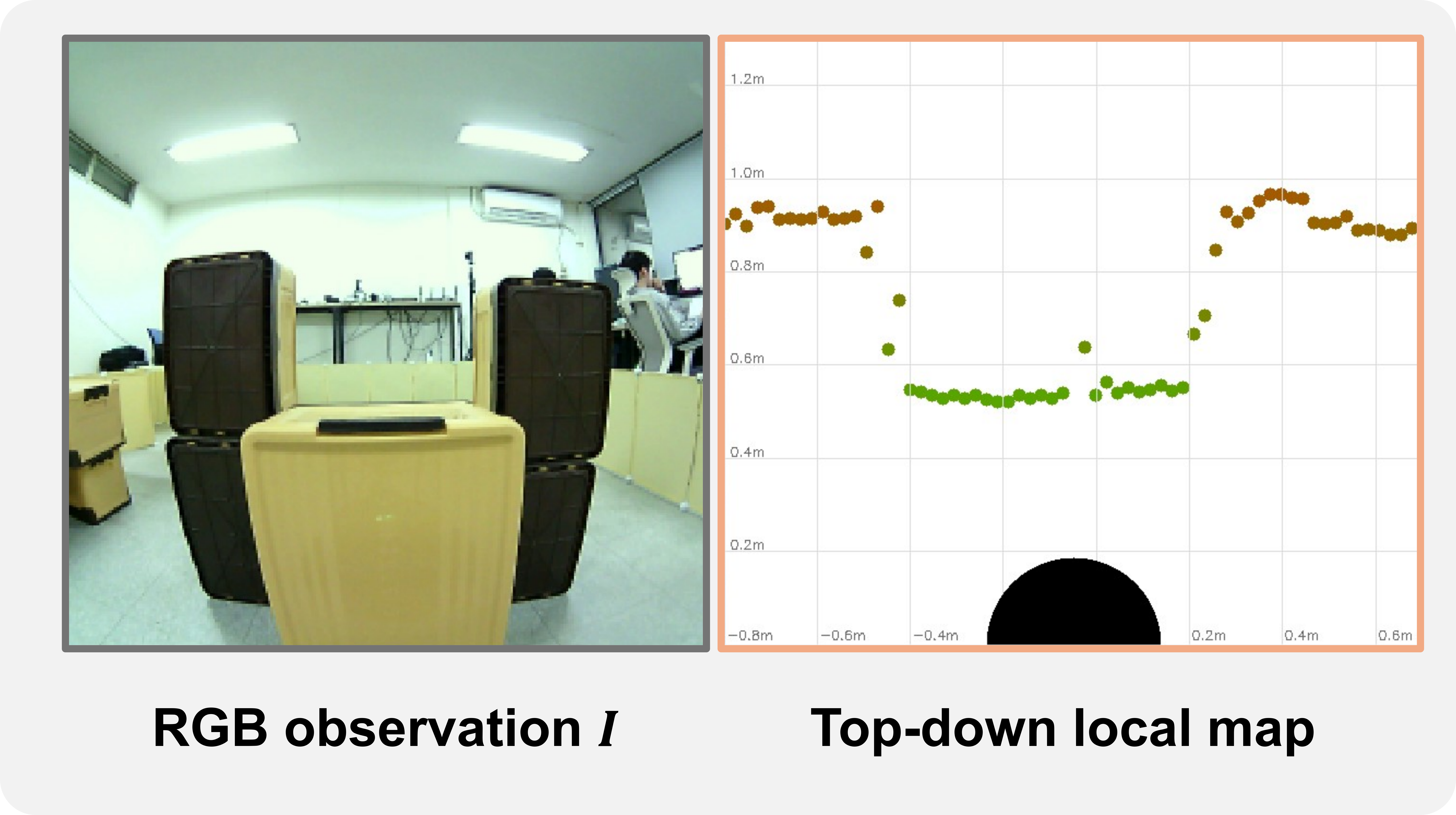}
        \caption{Top-down Depth Projection}
        \label{fig:top-down}
    \end{subfigure}%
    \hspace{0.2cm}
    \begin{subfigure}[t]{0.45\linewidth}
        \centering
        \includegraphics[width=\linewidth]{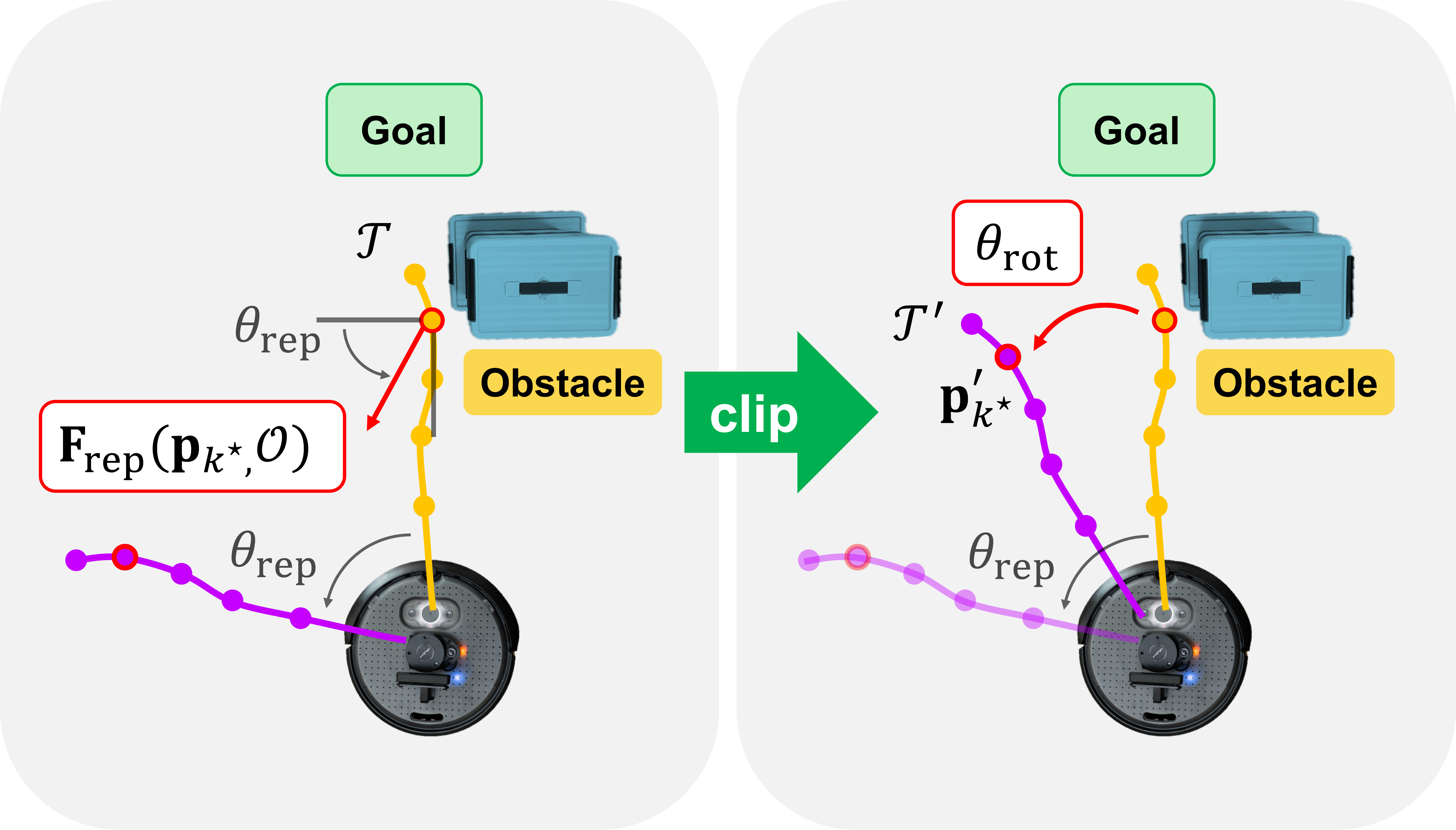}
        \caption{Repulsive Rotation Adjustment}
        \label{fig:traj_rot}
    \end{subfigure}
    \caption{(a) Top-down projection of estimated depth. Obstacle points are sampled every 10 pixels from the depth map (Left: input RGB, Right: top-down view), with colored dots indicating sampled points and the black circle denoting the robot. (b) Trajectory adjustment using repulsive force $\mathbf{F}_{\text{rep}}$. The original trajectory $\mathcal{T}$ (yellow) is rotated by $\theta_{\text{rep}}$, clipped to $\theta_{\text{clip}}$, resulting in the adjusted trajectory $\mathcal{T}'$ (purple).}
    \label{fig:rep_force}
    \vspace{-1em}
\end{figure}

Finally, \textit{RotateTrajectory}$(\mathcal{T}, \theta_{\text{rot}})$ (Algorithm~\ref{alg:care}, Line~\ref{alg:rotate}) applies a 2D rotation matrix $R \in \mathbb{R}^{2\times2}$ to each waypoint, yielding the adjusted trajectory $\mathcal{T}' = \{ \mathbf{p}'_1, \ldots, \mathbf{p}'_K \}$, where each waypoint is updated as $\mathbf{p}'_k = R(\theta_{\text{rot}}) \cdot \mathbf{p}_k$. This rotation steers the trajectory away from nearby obstacles while preserving its overall direction toward the goal (see Figure~\ref{fig:traj_rot}).
\vspace{-5pt}

\subsection{Safety-Enhancing Mechanism}
\label{sec:safe_enhance}
\vspace{-5pt}


Since vision-based models are limited to the field of view (FOV) of the camera, they cannot account for obstacles outside it. Sharp turns during forward motion can cause nearby obstacles to enter the view suddenly, giving the robot little time to react. To mitigate this risk, CARE introduces a Safe-FOV mechanism that suppresses forward motion when large heading changes are required.

First, we select $\mathbf{p'}_{k^\star} = (p'_{{k^\star}, x}, p'_{k^\star, y}) \in \mathcal{T}'$, the adjusted waypoint with the strongest repulsive response (as defined in Section~\ref{sec:repulsive}), as the basis for computing the desired heading. The desired heading $\theta_{\text{des}}$ is then computed using the function \textit{ComputeDesiredHeading}$(\mathcal{T}')$ (Algorithm~\ref{alg:care}, Line~\ref{alg:compute}). Since waypoints are generated relative to the local frame of the robot, $\theta_{\text{des}}$ can be directly calculated using the two-argument $\arctan2$ function $\theta_{\text{des}} = \arctan2(p'_{k^\star, y}, p'_{{k^\star}, x})$. We define a rotation threshold $\theta_{\text{thres}} \in \mathbb{R}$, above which forward motion is suppressed to prioritize in-place safe turns by $\textit{RotateInPlace}(\theta_{\text{des}})$. Otherwise, $\textit{MoveForwardAndTurn}(\theta_{\text{des}})$ is performed to move forward and turn simultaneously. In other words, the safety-enhancing mechanism (Lines~\ref{alg:safetybegin}--\ref{alg:safetyend} of Algorithm~\ref{alg:care}) follows
\begin{equation}
(v, \omega) =
    \begin{cases}
    (0, \omega_{\text{rot}}), & \text{if } |\theta_{\text{des}}| > \theta_{\text{thres}} \\
    (v_{\text{fwd}}, \omega_{\text{rot}}), & \text{otherwise}
    \end{cases}
\label{eq:safety_control}
\end{equation}
where $v_{\text{fwd}}$ denotes the forward linear velocity, and $\omega_{\text{rot}}$ is the angular velocity constrained by the maximum linear and angular velocity ($v_{\text{max}}, \omega_{\text{max}}$) . If the heading deviation $|\theta_{\text{des}}|$ exceeds the threshold, CARE commands in-place rotation until the deviation falls within $|\theta_{\text{des}}| \leq \theta_{\text{thres}}$, thereby reducing the risk of collisions incurred by the limited FOV of cameras, especially in dense environments.


In summary, CARE selectively adjusts trajectories based on obstacle presence. When nearby obstacles are detected via monocular depth estimation (i.e., within sensing range $\tau_z$), repulsive forces modify the trajectory. If no obstacles are found ($\mathcal{O} = \emptyset$), the original trajectory from the vision-based model is preserved. This design allows CARE to enhance safety while maintaining the generalization of pretrained models without fine-tuning or additional hardware.
\vspace{-0.5em}






\section{Experiments}
\label{sec:experiment}
\vspace{-1.0em}

We comprehensively evaluate CARE across unseen real environments and different robot platforms to validate our primary claim: CARE enhances collision avoidance capabilities of pretrained vision-based navigation models without fine-tuning, while preserving original navigation performance.
To validate the generalizability of CARE, we deploy our method on three mobile robot platforms with different camera configurations: (i) LoCoBot with a \ang{170} FOV fisheye camera\footnote{The camera setup is consistent with that used in the baseline models NoMaD and ViNT.}, (ii) Clearpath TurtleBot4 equipped with an \ang{89.5} FOV OAK-D Pro camera, and (iii) DJI RoboMaster S1 with a \ang{120} FOV camera. All robots operate under consistent velocity constraints (maximum linear velocity $v_{\text{max}}$ of $0.2\,\mathrm{m/s}$ and angular velocity 
$w_{\text{max}}$ of $0.8\,\mathrm{rad/s}$) for comparable evaluation. We optimize CARE parameters experimentally for each platform while maintaining consistent evaluation across all platforms. Detailed robot-specific parameters are provided in the Appendix.

\begin{figure}[t!]
\captionsetup{skip=0pt}
    \centering
    \begin{subfigure}[t]{0.45\linewidth}
        \centering
        \includegraphics[width=\linewidth]{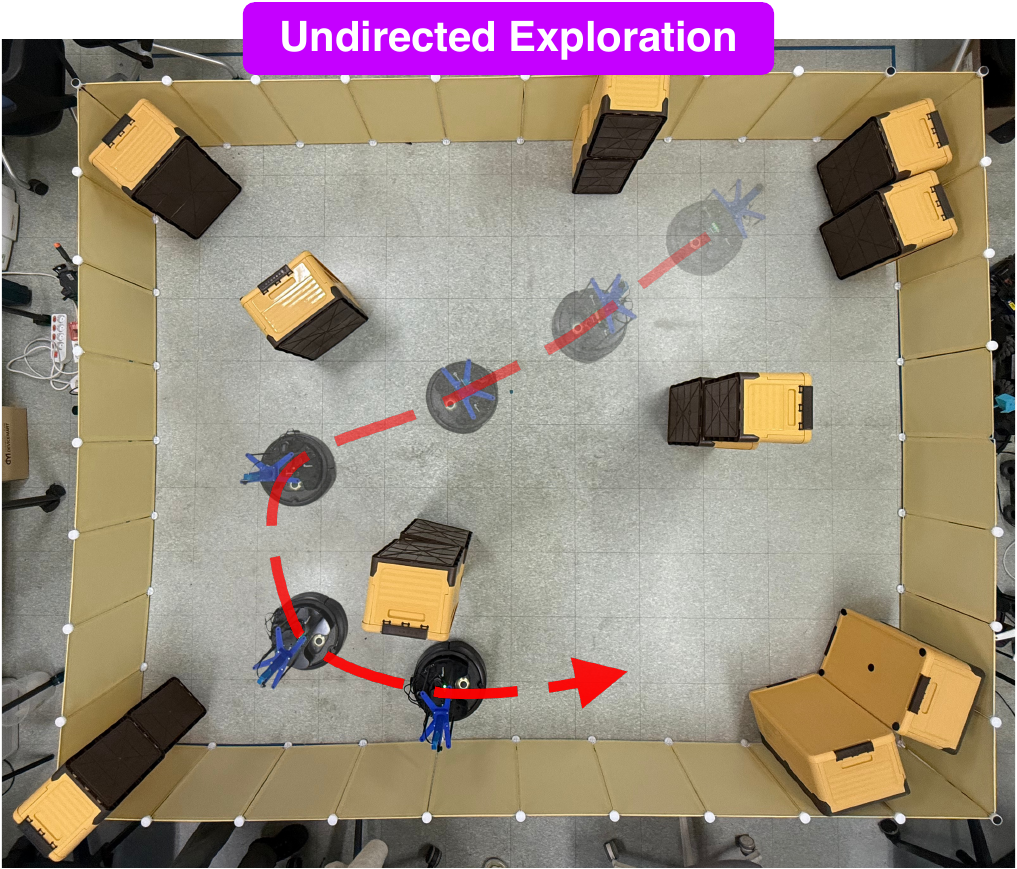}
        \caption{}
        \label{fig:exploration}
    \end{subfigure}
    \hspace{0.2cm}
    \begin{subfigure}[t]{0.45\linewidth}
        \centering
        \includegraphics[width=\linewidth]{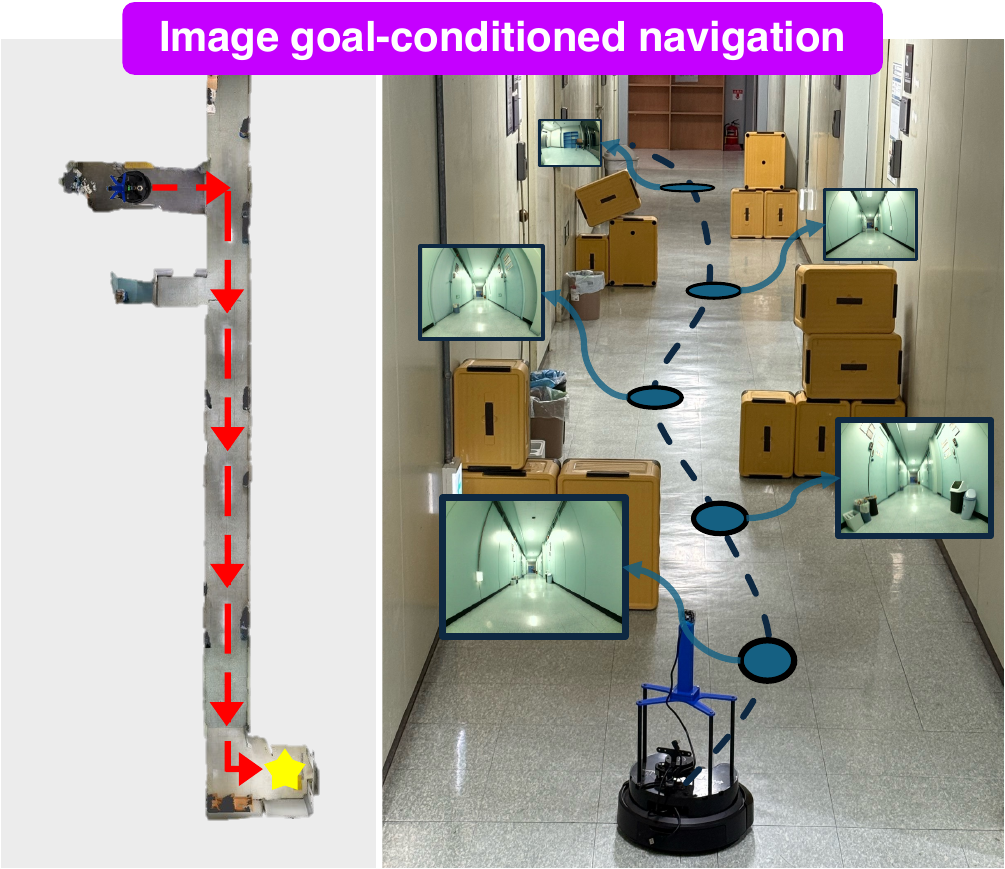}
        \caption{}
        \label{fig:navigation}
    \end{subfigure}
    \caption{Real-world experiments with CARE. (a) The unidirected exploration task in a confined environment ($3.5\,\mathrm{m} \times 2.8\,\mathrm{m}$) tests the ability to explore among unseen obstacles (boxes) as much as possible. (b) The image goal-conditioned navigation task in a corridor ($24\,\mathrm{m}$) tests the ability to reach the goal while avoiding seen obstacles (walls and doors) and unseen obstacles (boxes). The topological graph shows nodes with only image data.}
    \label{fig:environment}
    \vspace{-1.7em}
\end{figure}


\textbf{Undirected exploration:} This task evaluates the ability of the robot to explore safely in an environment with unseen objects. We place 10 OOD obstacles (plastic boxes) in a $3.5\,\mathrm{m} \times 2.8\,\mathrm{m}$ area as shown in the left of Figure~\ref{fig:exploration}. Since the robot explores without a destination, we measure the distance traveled until the first collision. To avoid indefinitely long trials, we stop if the robot travels above $30\,\mathrm{m}$. We compare NoMaD and CARE-integrated NoMaD.\footnote{We excluded ViNT from exploration experiments because, as a goal-conditioned model, it requires a goal image to generate trajectories and cannot operate without one. Supporting exploration would require a subgoal generator, typically implemented as a diffusion model trained in the target environment, which conflicts with our zero-shot assumption that prohibits additional training.} For each combination of robot platform and navigation method, we conduct 20 trials and report the average distance traveled. 

\textbf{Image goal-conditioned navigation:} This task tests the image goal-conditioned navigation capability of the robot in environments with random unseen obstacles (a $24\,\mathrm{m}$ corridor). Initially, a topological graph of corridors (see Figure~\ref{fig:navigation}) is constructed, and we place OOD obstacles at random locations along the corridor. We conducted 10 distinct setups, testing all robot and method combinations. The robot must navigate to the goal using the topological graph while avoiding obstacles as much as possible. Evaluation metrics include average path length (shorter is more efficient), completion time (faster is more efficient), and collision count (lower is safer). We also measure the arrival rate (higher is better), which is defined as the proportion of successful goal reaches, even if collisions occur. We compare NoMaD and ViNT and CARE-integrated NoMaD and ViNT.
\vspace{-0.5em}

\subsection{Undirected Exploration task in Unseen Environments} \vspace{-0.5em}

\begin{figure}[t]
\centering
\begin{minipage}{0.47\linewidth}
    \centering
    \includegraphics[width=\linewidth]{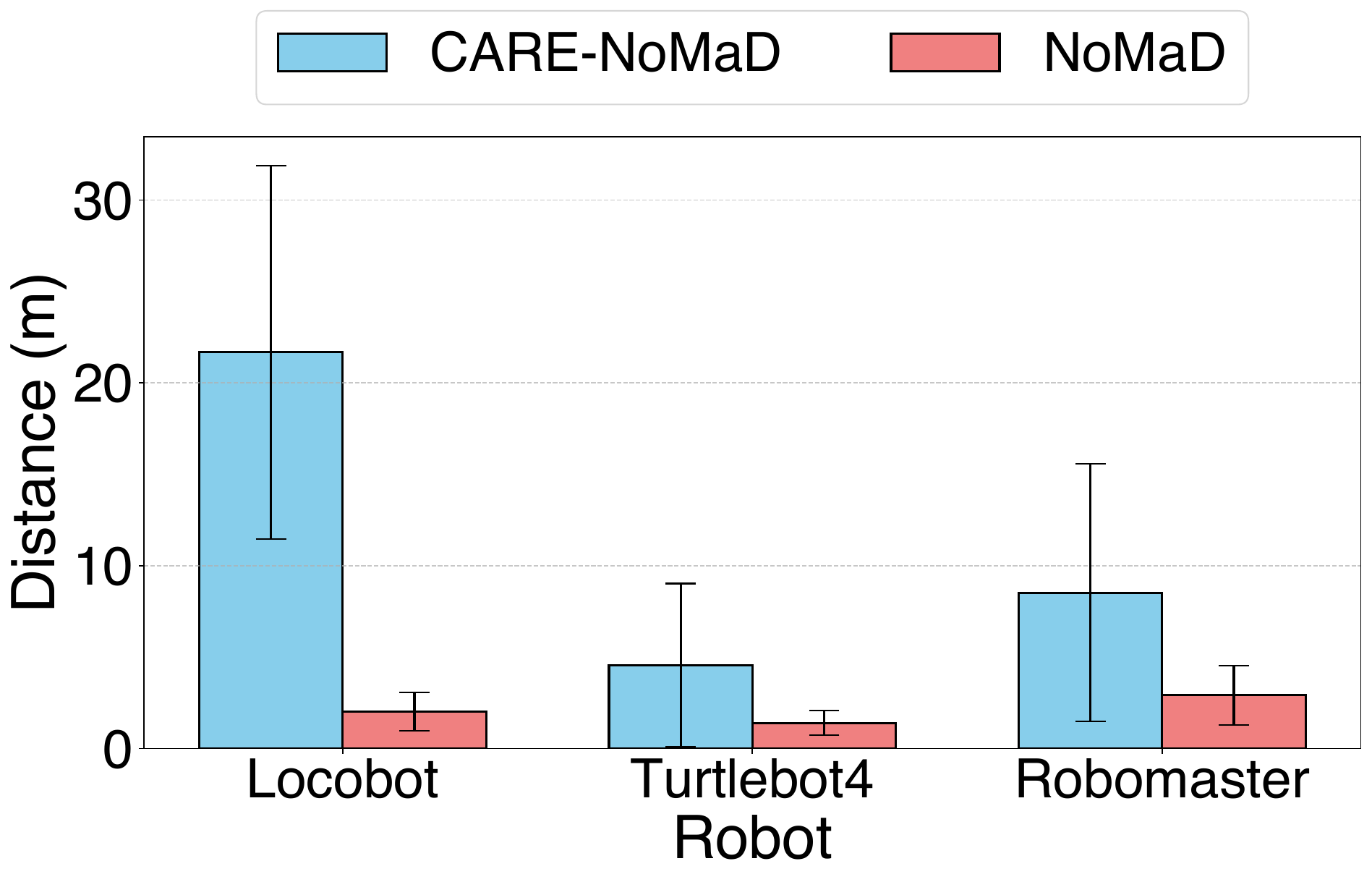}
    \vspace{-3ex} 
    \captionof{figure}{Comparison of distance traveled before first collision between CARE-NoMaD and baseline NoMaD across different robots.}
    \label{fig:exploration_experiment}
\end{minipage}
\hfill
\begin{minipage}{0.43\linewidth}
    \centering
    \renewcommand{\arraystretch}{1.2}
    \small
    \begin{tabular}{|l|l|l|}
    \hline
    Robot & Method & Distance (m) \\ \hline
    \multirow{2}{*}{Locobot} & CARE & $\textbf{21.67} \pm \textbf{10.20}$ \\ \cline{2-3}
     & NoMaD & $2.02 \pm 1.06$ \\ \hline
    \multirow{2}{*}{Turtlebot4} & CARE & $\textbf{4.55} \pm \textbf{4.47}$ \\ \cline{2-3}
     & NoMaD & $1.39 \pm 0.67$ \\ \hline
    \multirow{2}{*}{RoboMaster} & CARE & $\textbf{8.52} \pm \textbf{7.04}$ \\ \cline{2-3}
     & NoMaD & $2.91 \pm 1.63$ \\ \hline
    \end{tabular}
    \vspace{3ex} 
    \captionof{table}{Mean distance traveled with standard deviation before collision for each robot platform and navigation method.}
    \label{tab:exploration}
\end{minipage}
\vspace{-10pt}
\end{figure}

Figure~\ref{fig:exploration_experiment} and Table~\ref{tab:exploration} present our exploration experiment results. Across all settings, CARE enables robots to travel $2.9\times$ to $10.7\times$ longer before the first collision compared to NoMaD and ViNT without CARE. These performance variations correlated with the camera specifications of each configuration and the resulting depth estimation quality.

\textbf{LoCoBot} with CARE exhibits both the longest distance traveled and the most significant improvement ($10.7\times$). This performance increase can be attributed to two main factors: the wide-angle fisheye camera enables extensive environmental perception, effectively detecting peripheral obstacles; and NoMaD, which has been predominantly trained on fisheye camera data, performs better with the camera setup of LoCoBot, which CARE further enhances.


\textbf{TurtleBot4} with CARE exhibits moderate improvement ($3.3\times$) despite having the lowest distance traveled. This enhancement is partly due to its camera mounting position. When navigating corners, the rear-mounted camera maintains visibility of obstacles throughout the turn, allowing CARE to make safer adjustments. However, its narrow FOV still limits overall performance, particularly when approaching walls head-on.


\textbf{RoboMaster} without CARE achieves higher distance traveled due to its wider FOV, but shows the smallest improvement ($2.9\times$) with CARE. This limited gain stems from its front-mounted camera. During turning, the camera moves past obstacles before the body of the robot completes the turn, creating blind spots where side collisions can occur without detection. While the wider FOV achieves relatively higher distance traveled, it cannot fully compensate for these cornering blind spots.
\vspace{-3pt}




\subsection{Image Goal-Conditioned Navigation with Unseen Obstacles} 
\vspace{-0.7em}
\label{subsec:goal_navigation}
We evaluate NoMaD and ViNT and their CARE-integrated variants in image goal-conditioned navigation scenarios involving unseen obstacles with the three robot platforms. We first establish a topological graph in a corridor environment to enable global planning. During testing, we place several OOD obstacles not represented in the topological map, creating scenarios that would typically require model fine-tuning to handle appropriately. 
Table~\ref{tab:performance_comparison} presents comprehensive results comparing the baseline navigation models (NoMaD and ViNT) against the same models integrated with CARE. 
All experiments are conducted across 10 distinct obstacle settings, with randomized positions for each trial.\footnote{Detailed information about obstacle placements for each trial is available in Appendix~\ref{sec:test_instances}.}

\begin{table*}[t]
\centering
\small  
\setlength{\tabcolsep}{2pt}  
\caption{The comparison of the navigation performance between baseline vision-based models and their CARE-integrated counterparts across three robot platforms. Metrics include path length (m), completion time (s), collision counts, and goal arrival rate (\%).}
\label{tab:performance_comparison}
\scalebox{0.86}{
    \begin{tabular}{@{}llcccccccc@{}}
    \toprule
    \multirow{2}{*}{\textbf{Robot}} & \multirow{2}{*}{\textbf{Model}} & \multicolumn{4}{c}{\textbf{Vision-based Model}} & \multicolumn{4}{c}{\textbf{CARE-integrated Vision-based Model}} \\
    \cmidrule(lr){3-6} \cmidrule(lr){7-10}
     & & \textbf{Length} & \textbf{Time} & \textbf{\# Collision} & \textbf{Arrival} & \textbf{Length} & \textbf{Time} & \textbf{\# Collision} & \textbf{Arrival} \\
     & & \textbf{(m)} & \textbf{(s)} & & \textbf{Rate} & \textbf{(m)} & \textbf{(s)} & & \textbf{Rate} \\
    \midrule
    \multirow{2}{*}{Locobot} 
     & NoMaD & $26.39 \pm 1.02$ & $132.98 \pm 5.02$ & 0.7 & 50\% & $26.06 \pm 0.30$ & $136.06 \pm 3.90$ & \textbf{0.0} & \textbf{90\%} \\
     & ViNT & $25.79 \pm 0.19$ & $129.98 \pm 0.85$ & 0.4 & 50\% & $25.88 \pm 0.18$ & $134.87 \pm 2.40$ & \textbf{0.0} & \textbf{80\%} \\
    \midrule
    \multirow{2}{*}{Turtlebot4} 
     & NoMaD & $25.27 \pm 0.18$ & $127.28 \pm 0.55$ & 0.7 & 20\% & $26.35 \pm 1.10$ & $149.13 \pm 9.03$ & \textbf{0.3} & \textbf{50\%} \\
     & ViNT & $26.42 \pm 0.75$ & $133.73 \pm 3.82$ & 0.2 & 70\% & $26.55 \pm 0.98$ & $145.91 \pm 14.47$ & \textbf{0.1} & \textbf{100\%} \\
    \midrule
    \multirow{2}{*}{RoboMaster} 
     & NoMaD & $25.24 \pm 0.61$ & $126.33 \pm 3.05$ & 0.8 & 30\% & $25.89 \pm 0.96$ & $146.76 \pm 10.34$ & \textbf{0.1} & \textbf{80\%} \\
     & ViNT & $24.79 \pm 0.48$ & $125.60 \pm 2.28$ & 1.0 & 50\% & $25.75 \pm 1.10$ & $155.60 \pm 11.35$ & \textbf{0.2} & \textbf{80\%} \\
    \bottomrule
    \end{tabular}
}
\end{table*}
\vspace{-6pt}


\textbf{Path length and time:} The integration of CARE results in slightly longer paths (up to 4.27\% increase) and completion times (up to 23.89\% increase). This can be attributed to the adjusted trajectories around detected obstacles for safer navigation.
Despite these inefficiencies, the overall navigation performance remained reasonable.

\textbf{Collisions:} CARE consistently reduces collisions throughout all platforms. Most notably, LoCoBot achieves complete collision elimination with both NoMaD and ViNT models. For RoboMaster, CARE integration yields notable improvements (80--88\% reduction), while TurtleBot4 demonstrates modest but meaningful reductions (50--57\%).
Despite these improvements, CARE does not completely eliminate all collisions in the cases of TurtleBot4 and RoboMaster. One reason is that monocular depth estimation occasionally produces inaccurate results, particularly on obstacle surfaces without patterns or in challenging lighting conditions. 
Another reason is the limited FOV or mounting position of the camera, which causes obstacles outside the view to lead to side collisions with the robot body.



\textbf{Arrival rate:} The integration of CARE significantly enhances arrival rates across all platforms. On LoCoBot and RoboMaster, both NoMaD and ViNT models with CARE show substantial improvements. For TurtleBot4, ViNT already performs well (70\% baseline) and reaches perfect goal-reaching (100\%) with CARE, while NoMaD struggles even with safety enhancements, improving from only 20\% to 50\%. This shows that while CARE consistently adjusts trajectories to avoid collisions, the quality of trajectories generated by the base navigation model significantly influences results. Nevertheless, even with NoMaD on TurtleBot4, CARE still provides considerable improvement over the baseline model. Overall, CARE improves goal arrival reliability across different robots and camera configurations without requiring additional fine-tuning or sensor hardware.

\textbf{Navigation with dynamic obstacles.} We conducted additional tests in the same environment as the goal-conditioned experiments (see Figure~\ref{fig:navigation}), where up to three people abruptly crossed the robot path during navigation. A trial was considered successful if the robot reached the goal without collision. Unlike prior works, we introduced dynamic obstacles that appeared suddenly from outside the camera field of view. NoMaD often failed to react, while CARE-integrated robots consistently exhibited reactive stopping and safe avoidance, demonstrating the ability to handle unpredictable dynamic obstacles where vision-only models without fine-tuning often fail to generalize. Further details are provided in Appendix~\ref{sec:dynamic_obstacles}.

These results demonstrate the consistent ability of CARE to enhance navigation safety across various robot platforms without additional training. While performance varies based on camera configuration, CARE considerably improves collision avoidance in all tested scenarios, augmenting the capabilities of baseline models to handle complex environments with OOD objects. Supplementary videos of these experiments are available on our project page: \href{https://airlab-sogang.github.io/CARE/}{\textcolor{black}{https://airlab-sogang.github.io/CARE/}}
\vspace{-15pt}




\section{Conclusion}
\label{sec:conclusion}
\vspace{-1em}
We presented CARE, an attachable module that enhances the safety of vision-based navigation models without requiring fine-tuning or additional range sensors. By combining monocular depth estimation with repulsive force-based trajectory adjustment, CARE significantly improves collision avoidance in out-of-distribution environments and across diverse robot platforms with varying camera configurations. CARE enables zero-shot deployment of vision-based models in unseen environments where fine-tuning would otherwise be necessary, overcoming their limited collision avoidance capabilities. Extensive real-world experiments demonstrate that CARE substantially reduces collision rates and improves task success, while preserving the strong generalization performance and efficiency of the underlying navigation models. Our results highlight CARE as a simple and effective solution for safer real-world deployment of visual navigation systems.



\clearpage
\section{Limitations}
\label{sec:limitation}



Although CARE does not require fine-tuning of the navigation policy or additional range sensors, it has several limitations arising from its reliance on monocular depth estimation and vision-based navigation models.

First, CARE is sensitive to errors in monocular depth prediction. Reflective or transparent surfaces often lead to inaccurate depth values, which may cause the system to underestimate obstacle proximity or miss obstacles entirely. Additionally, depth estimation models may misclassify ground or low-texture areas, resulting in false avoidance maneuvers that deviate unnecessarily from the desired path.

Second, CARE only reacts to obstacles within a limited sensing range $\tau_z$ derived from the estimated depth. Distant obstacles beyond this range are ignored, and the system cannot anticipate them. Furthermore, as with most vision-based systems, perception range of robot is restricted to the FOV of attached camera, making it unable to respond to dynamic obstacles approaching from behind or from occluded regions.

Third, being entirely vision-based, CARE inherits the limitations of RGB-only systems. Its performance may degrade under poor lighting conditions or in environments where camera orientation fails to capture critical obstacles. These constraints reduce the effectiveness of the repulsive adjustment in unpredictable or low-visibility scenarios.

Fourth, CARE adjusts but does not generate trajectories. Its effectiveness depends on the quality of the underlying navigation model. If the base policy predicts unsafe paths (such as directly toward an obstacle) CARE may be unable to fully correct them, even with strong repulsive forces. This limits its performance in densely cluttered or highly dynamic environments.

Finally, CARE introduces computational overhead from the depth estimation step. However, the model we employed (UniDepthV2) is lightweight and uses only 100MB of GPU memory, which does not impact real-time performance.

\section{Acknowledgments}
\label{sec:ack}

This work was supported by the National Research Foundation of Korea (NRF) grant funded by the Korea government (MSIT) (No. RS-2024-00461409), Korea Planning \& Evaluation Institute of Industrial Technology (KEIT) grant funded by the Korea government (MOTIE) (RS-2024-00444344), and Institute of Information \& communications Technology Planning \& Evaluation (IITP) grant funded by the Korea government(MSIT) (RS-2022-00143911, AI Excellence Global Innovative Leader Education Program).



\bibliography{example}  

\clearpage
\appendix
\section*{\LARGE{Appendix}}


\textbf{A}\quad \textbf{\hyperref[sec:add_exp]{Additional Experiments}} \dotfill \textbf{\pageref{sec:add_exp}} \\
A.1\quad \hyperref[sec:dynamic_obstacles]{Dynamic Obstacles in Goal-Conditioned Navigation} \dotfill \pageref{sec:dynamic_obstacles} \\
A.2\quad \hyperref[sec:seen_env]{Performance in Seen Environments} \dotfill \pageref{sec:seen_env} \\

\textbf{B}\quad \textbf{\hyperref[sec:exp_details]{Experimental Details}} \dotfill \textbf{\pageref{sec:exp_details}} \\
B.1\quad \hyperref[sec:test_instances]{Test Instances for Goal-Conditioned Navigation} \dotfill \pageref{sec:test_instances} \\
B.2\quad \hyperref[sec:robot_specs]{Robot Platforms and Specifications} \dotfill \pageref{sec:robot_specs} \\
B.3\quad \hyperref[sec:care_params]{Common Parameters for CARE} \dotfill \pageref{sec:care_params} \\
B.4\quad \hyperref[sec:depth_model]{Depth Estimation Model: UniDepthV2} \dotfill \pageref{sec:depth_model} \\
B.5\quad \hyperref[sec:traj_selection]{Trajectory Selection in NoMaD and ViNT} \dotfill \pageref{sec:traj_selection} \\

\vspace{1em}

\section{Additional Experiments}
\label{sec:add_exp}

\subsection{Dynamic Obstacles in Goal-Conditioned Navigation}
\label{sec:dynamic_obstacles}

To assess the robustness of CARE under dynamic conditions, we evaluate the system in an image goal-conditioned navigation setting with abrupt human intervention. Unlike the evaluations in the paper, this setup contains no static out-of-distribution (OOD) obstacles. All tests are conducted using the LoCoBot platform.

\textbf{Dynamic Obstacle Scenarios.} We introduce up to three humans acting as dynamic, unexpected obstacles, designed to test reactive collision avoidance. Each test trial includes one of the following scenarios\footnote{Please refer to the supplementary video for better understanding.}.
\begin{enumerate}[label=(\roman*), leftmargin=*]
    \item A person appears from a side outside the field of view.
    \item A person appears from behind and overtakes the robot, stopping in front.
    \item A person walks toward the robot from the front and stops directly in its path.
\end{enumerate}

\textbf{Metrics and setup.} For each of the three human intervention types, we run 10 trials with four methods: NoMaD, NoMaD+CARE, ViNT, and ViNT+CARE. We record the number trials where collisions occur.

\begin{table}[h]
\centering
\caption{Number trials where collisions occur (out of 10 trials) for each dynamic obstacle type}
\label{tab:dynamic_results}
\begin{tabular}{lccc}
\toprule
\textbf{Model} & \textbf{(i)  Corner-appear} & \textbf{(ii) Behind-to-front} & \textbf{(iii) Front-approach} \\
\midrule
NoMaD & 8/10 & 10/10 & 10/10 \\
NoMaD + CARE & \textbf{0/10} & \textbf{0/10} & \textbf{0/10} \\
ViNT & 7/10 & 10/10 & 10/10  \\
ViNT + CARE & \textbf{0/10} & \textbf{0/10} & \textbf{0/10} \\
\bottomrule
\end{tabular}
\end{table}

\textbf{Results and Analysis.}
Table~\ref{tab:dynamic_results} shows the number of collisions in each dynamic obstacle scenario. Without CARE, both NoMaD and ViNT fail to react effectively to abrupt human interventions, resulting in 7 to 10 collisions. In contrast, CARE-integrated policies complete all trials without a single collision, where this improvement is attributed to the repulsive force estimation based on predicted depth, enabling reliable detection of human legs even under sudden appearances. 

In scenario \textbf{(i)}, where a person enters from the side, both NoMaD and ViNT occasionally succeed in generating avoidance waypoints in the opposite direction, thereby avoiding some collisions. However, in scenarios \textbf{(ii)} and \textbf{(iii)}, where the human appears directly in front of the robot, both models fail in all trials. The key reason is the lack of geometric awareness in NoMaD and ViNT: although the models sometimes attempt avoidance by turning slightly, the generated waypoints do not ensure sufficient clearance. As a result, the robots consistently collide after small heading changes, or fail to respond to sudden frontal appearances.

In contrast, CARE-enabled policies successfully avoid all collisions in these scenarios. The estimated depth information allows CARE to compute repulsive directions with adequate lateral displacement, while the Safe-FOV mechanism suppresses forward motion until the robot achieves a sufficiently safe heading. This combination enables reliable and robust collision avoidance even under abrupt dynamic disturbances.


\subsection{Performance in Seen Environments}
\label{sec:seen_env}

We evaluate NoMaD and ViNT in a fully seen environment without added OOD obstacles as a sanity check to verify that our implementations reproduce the original models. All experiments are conducted using the LoCoBot platform.

\textbf{Setup.} The test environment and overall procedure follow the same setup as in the goal-conditioned navigation experiments in our paper. A topological graph is predefined using a sequence of image goals. For fair comparison, the subgoal image sequences fed to both NoMaD and ViNT are fixed and reused across all trials. This sequence is also identical to those used in the OOD experiments presented in the main paper.

\textbf{Metric.} We measure the success rate, defined as the percentage of trials in which the robot reaches the image goal without any collisions. Each configuration is tested for 10 trials.

\begin{table}[h]
\centering
\caption{Success rate in seen environments without OOD obstacles}
\label{tab:seen_env_results}
\begin{tabular}{lc}
\toprule
\textbf{Model} & \textbf{Success Rate (10 trials)} \\
\midrule
NoMaD & 100\% \\
ViNT  & 100\% \\
\bottomrule
\end{tabular}
\end{table}

\textbf{Results and Analysis.} 
Both NoMaD and ViNT achieve a 100\% success rate in the seen environment, completing all 10 trials without collision or failure. These results confirm that both the pretrained vision-based models and the constructed topological navigation pipeline function reliably under seen conditions. 

The experiments are conducted on the LoCoBot platform (see Sec.~\ref{sec:robot_specs}), which closely matches the setup used in the original NoMaD and ViNT papers and is also employed to train and tune the released pretrained weights\footnote{https://github.com/robodhruv/visualnav-transformer}. The consistent performance thus validates the correct integration and implementation of the models. This outcome supports the analysis that the performance degradation observed in other experiments (e.g., with OOD obstacles or dynamic human interventions) is not due to flaws in the experimental setup or execution, but rather reveals the inherent limitations of vision-only navigation models when deployed in unfamiliar environments without fine-tuning.

\section{Experimental Details}
\label{sec:exp_details}


\subsection{Test Instances for Goal-Conditioned Navigation}
\label{sec:test_instances}

\begin{figure*}[h]
    \centering
    \includegraphics[width=0.195\linewidth]{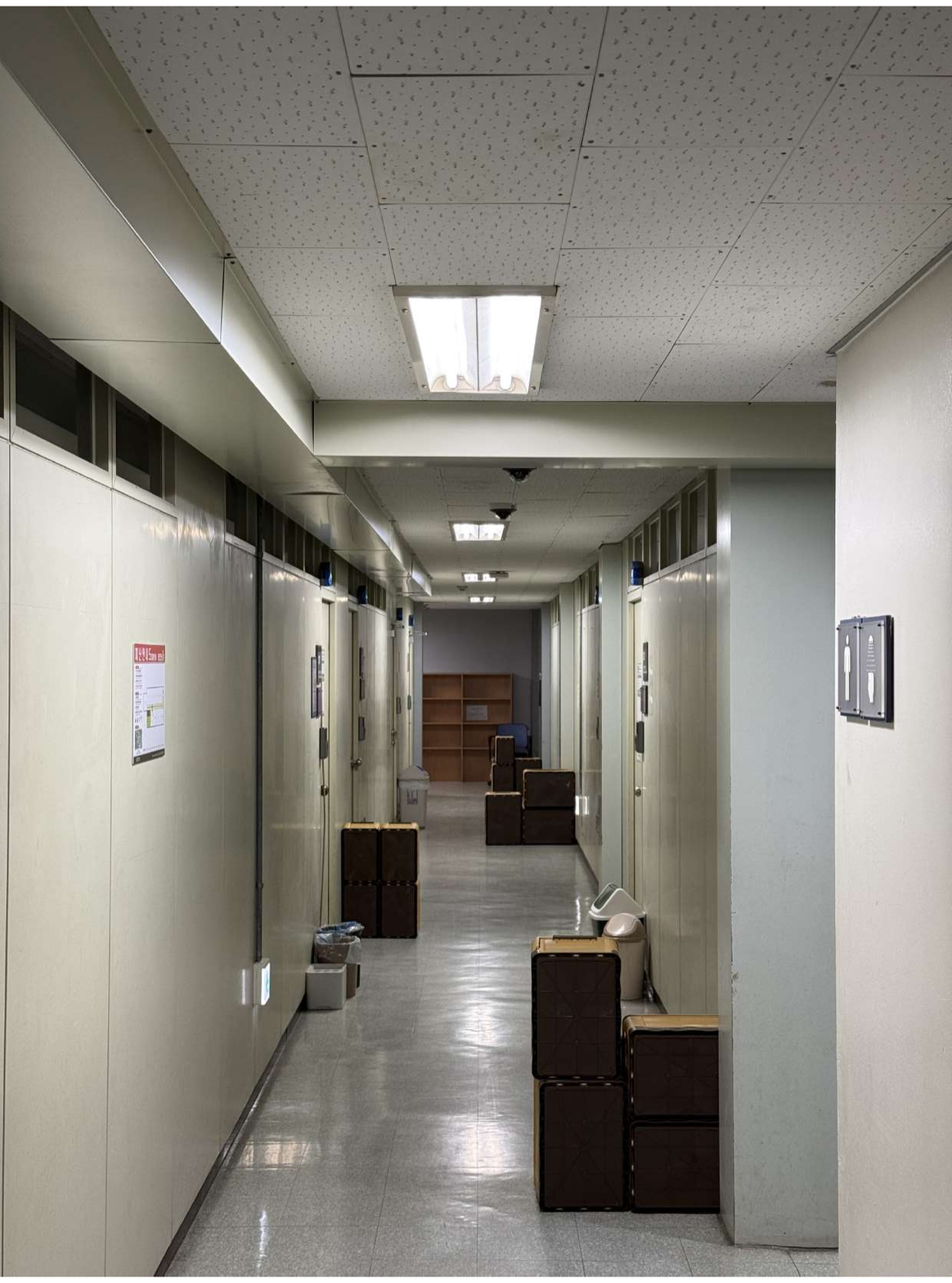}
    \includegraphics[width=0.195\linewidth]{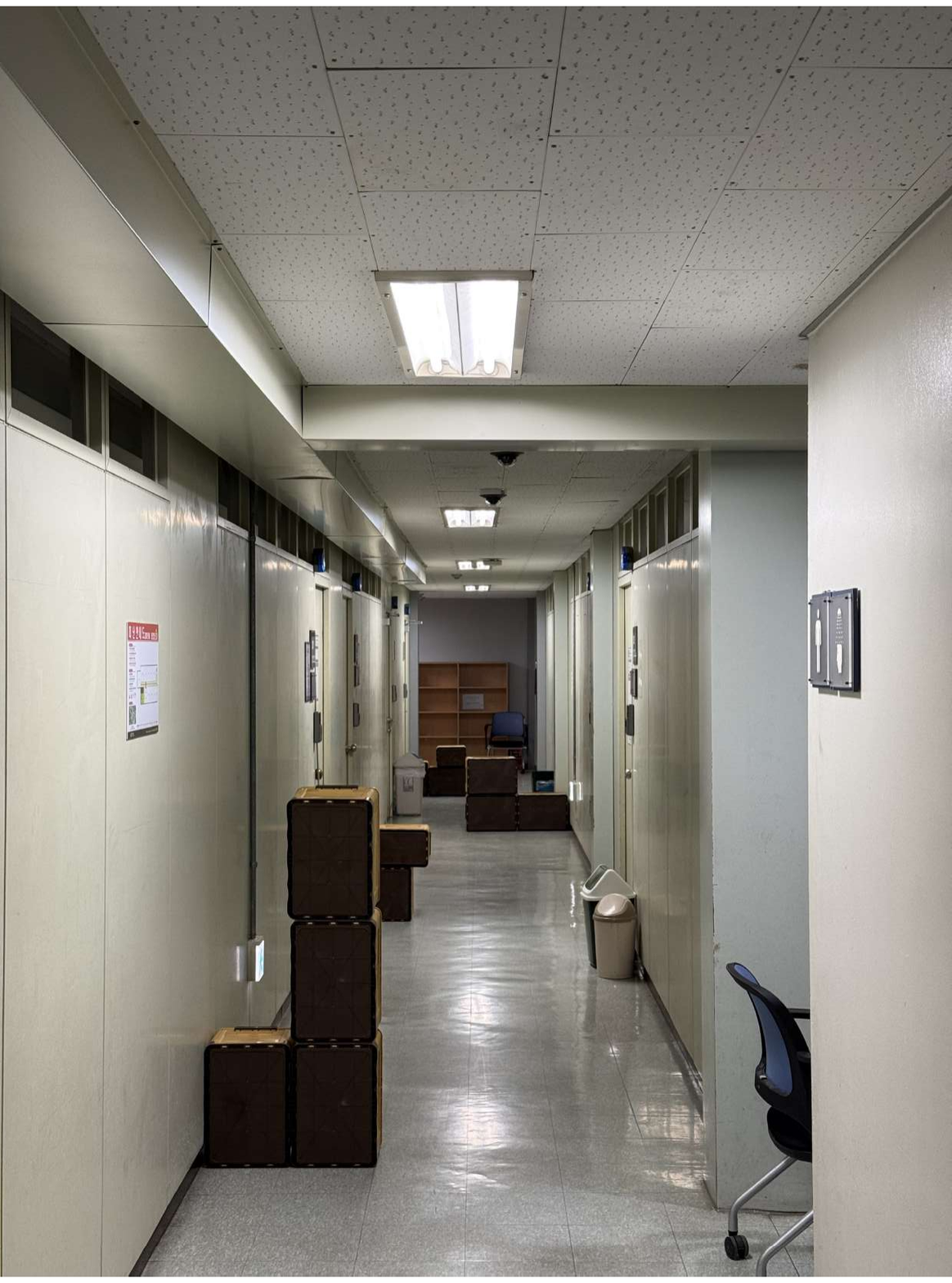}
    \includegraphics[width=0.195\linewidth]{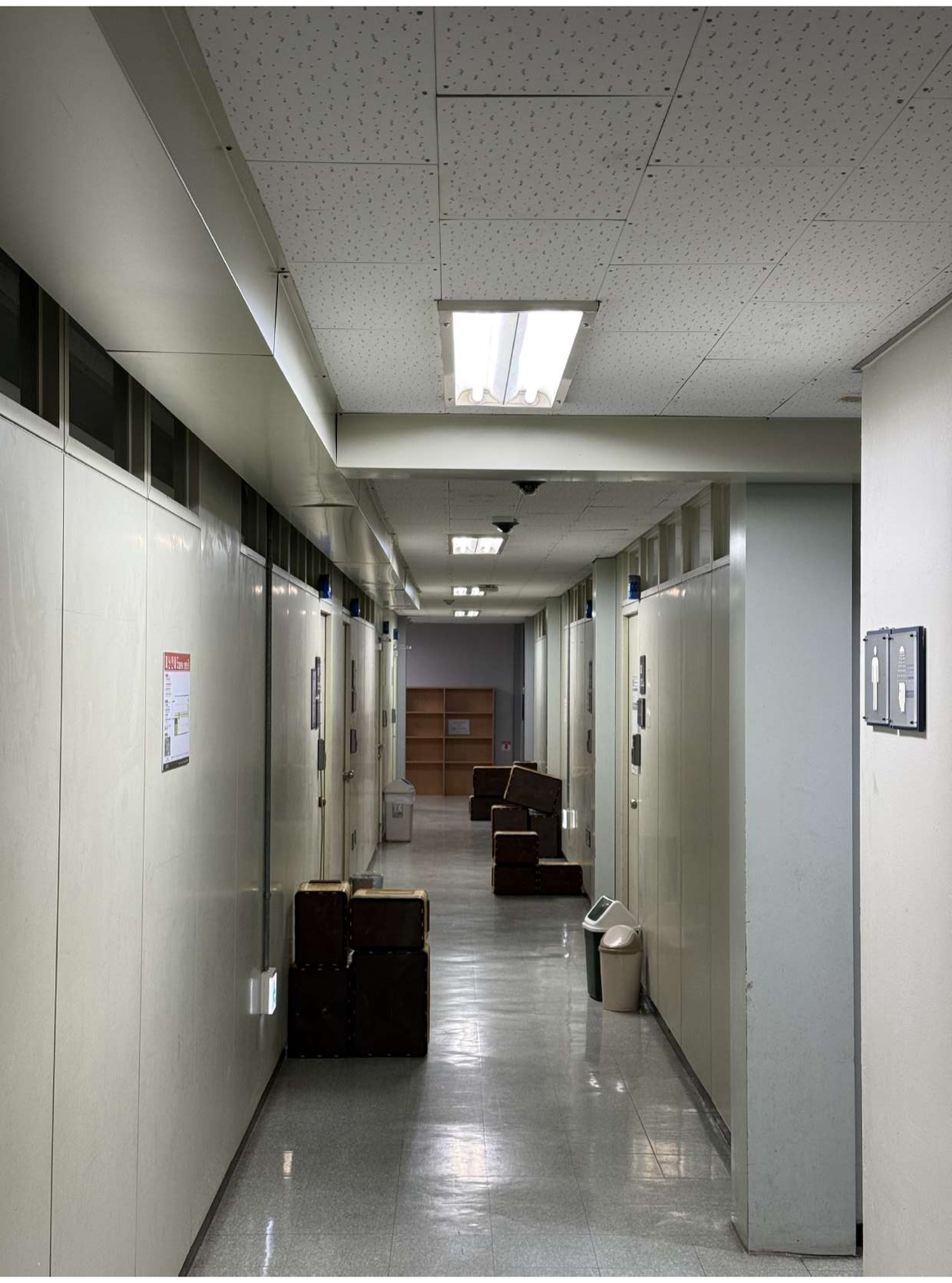}
    \includegraphics[width=0.195\linewidth]{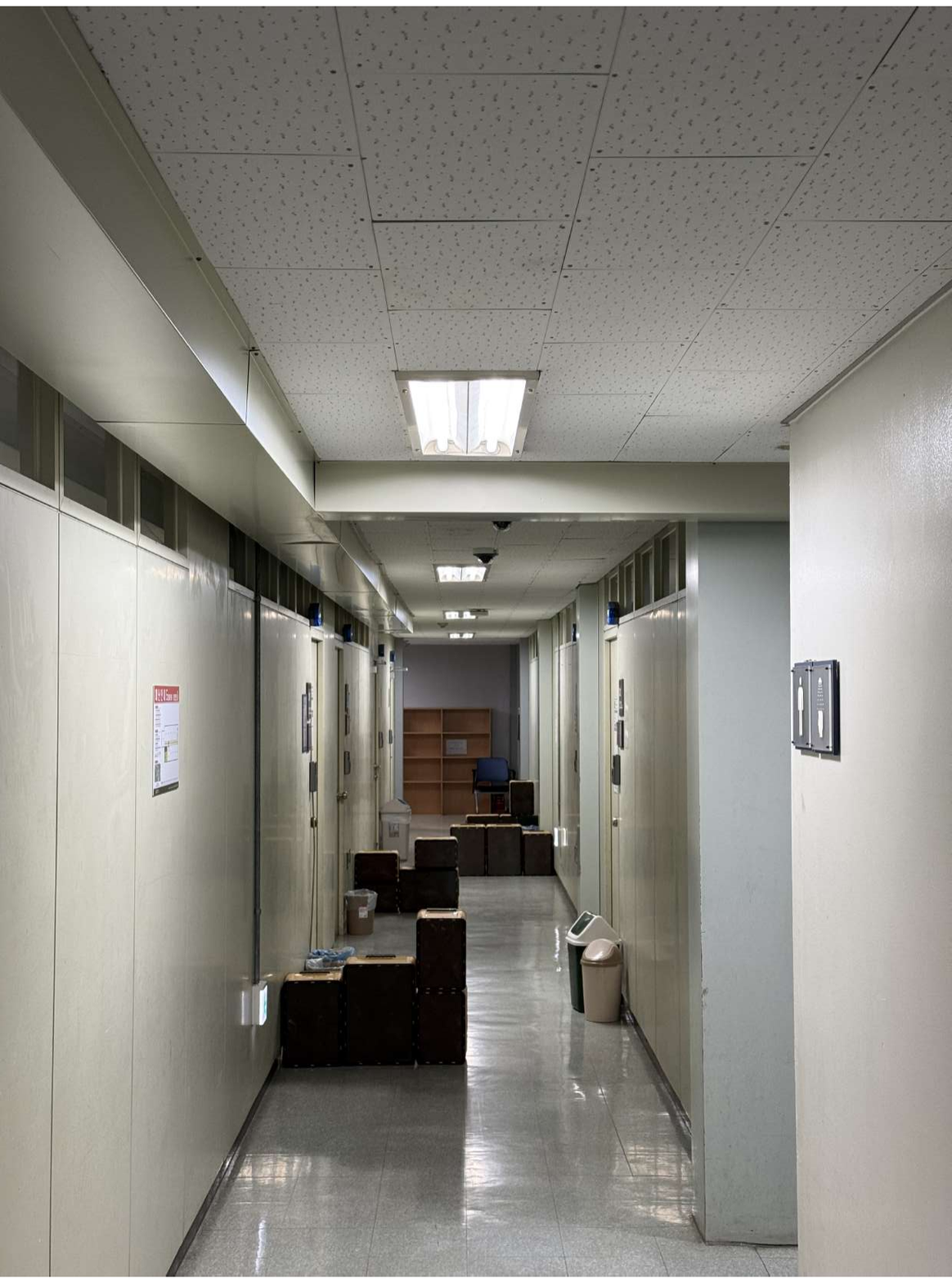}
    \includegraphics[width=0.195\linewidth]{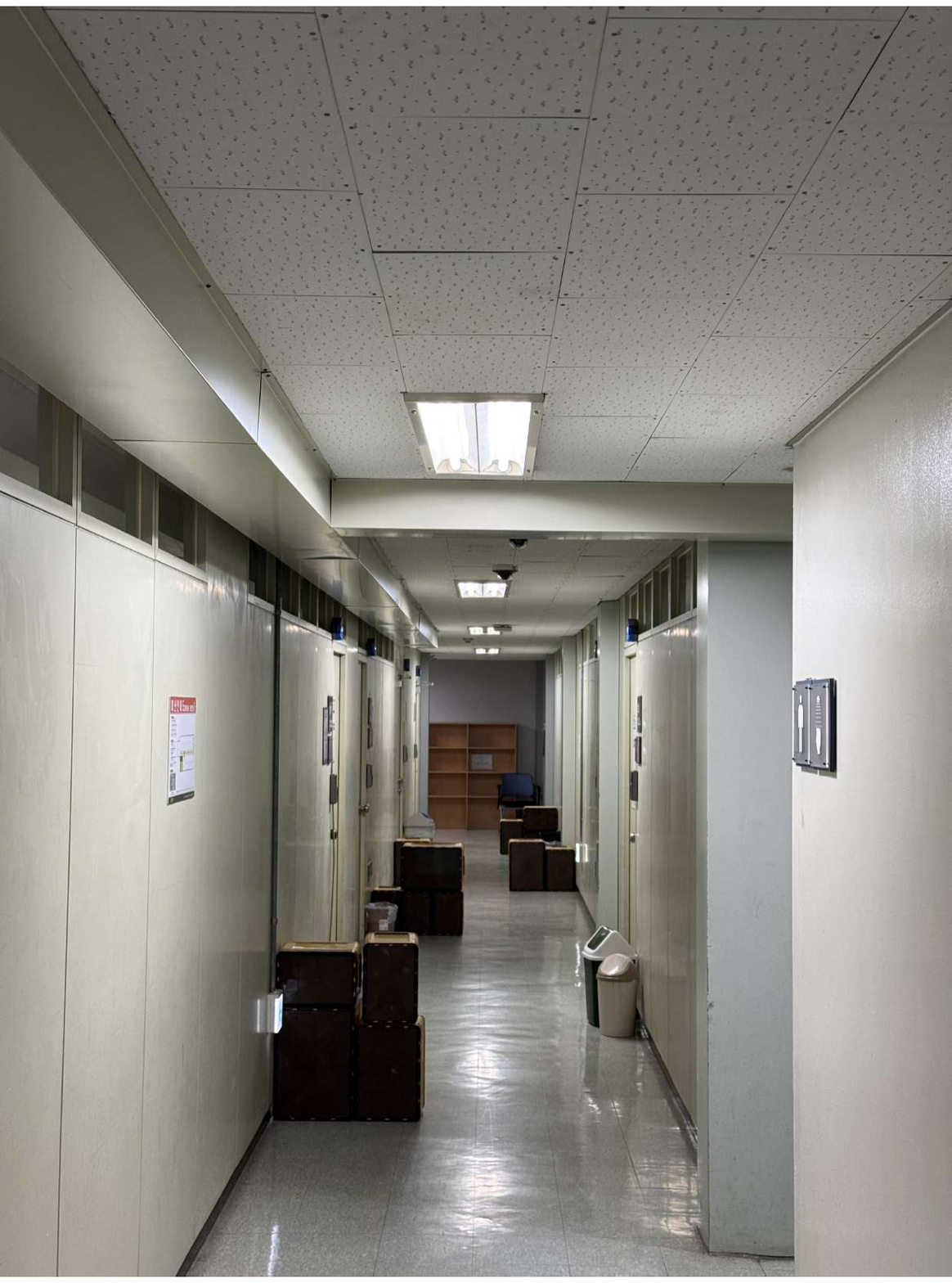}

    \vspace{3pt}
    \includegraphics[width=0.195\linewidth]{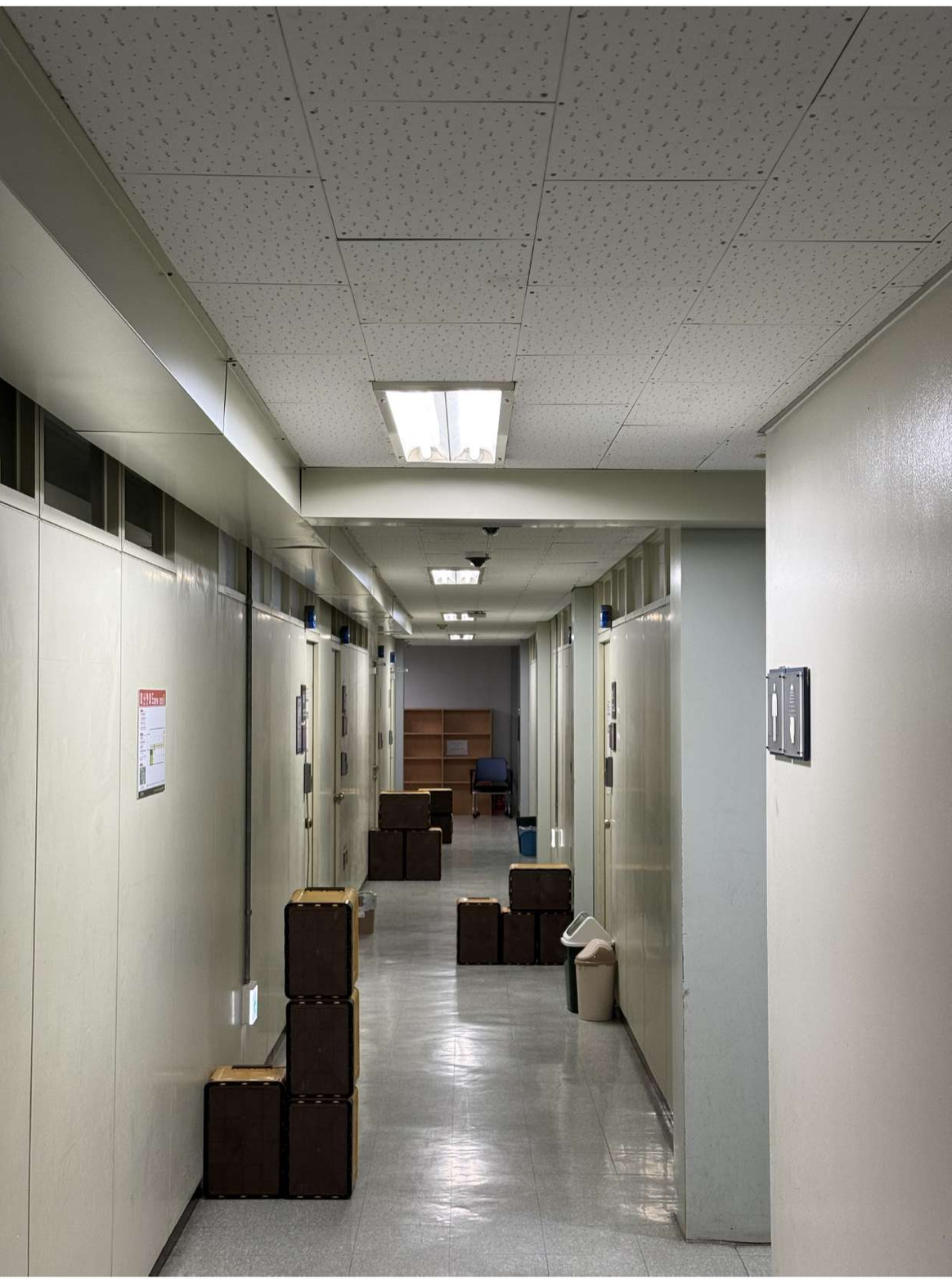}
    \includegraphics[width=0.195\linewidth]{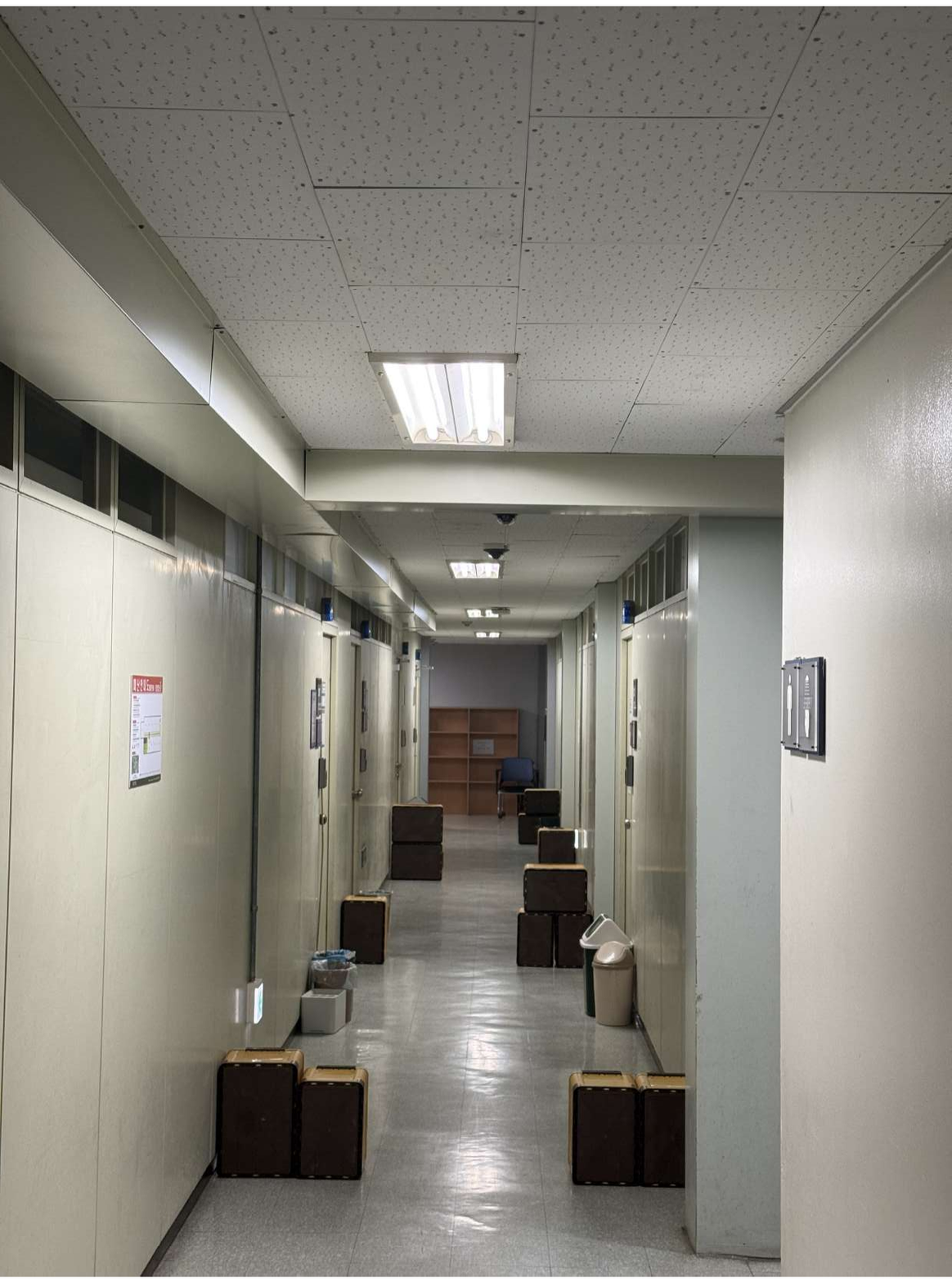}
    \includegraphics[width=0.195\linewidth]{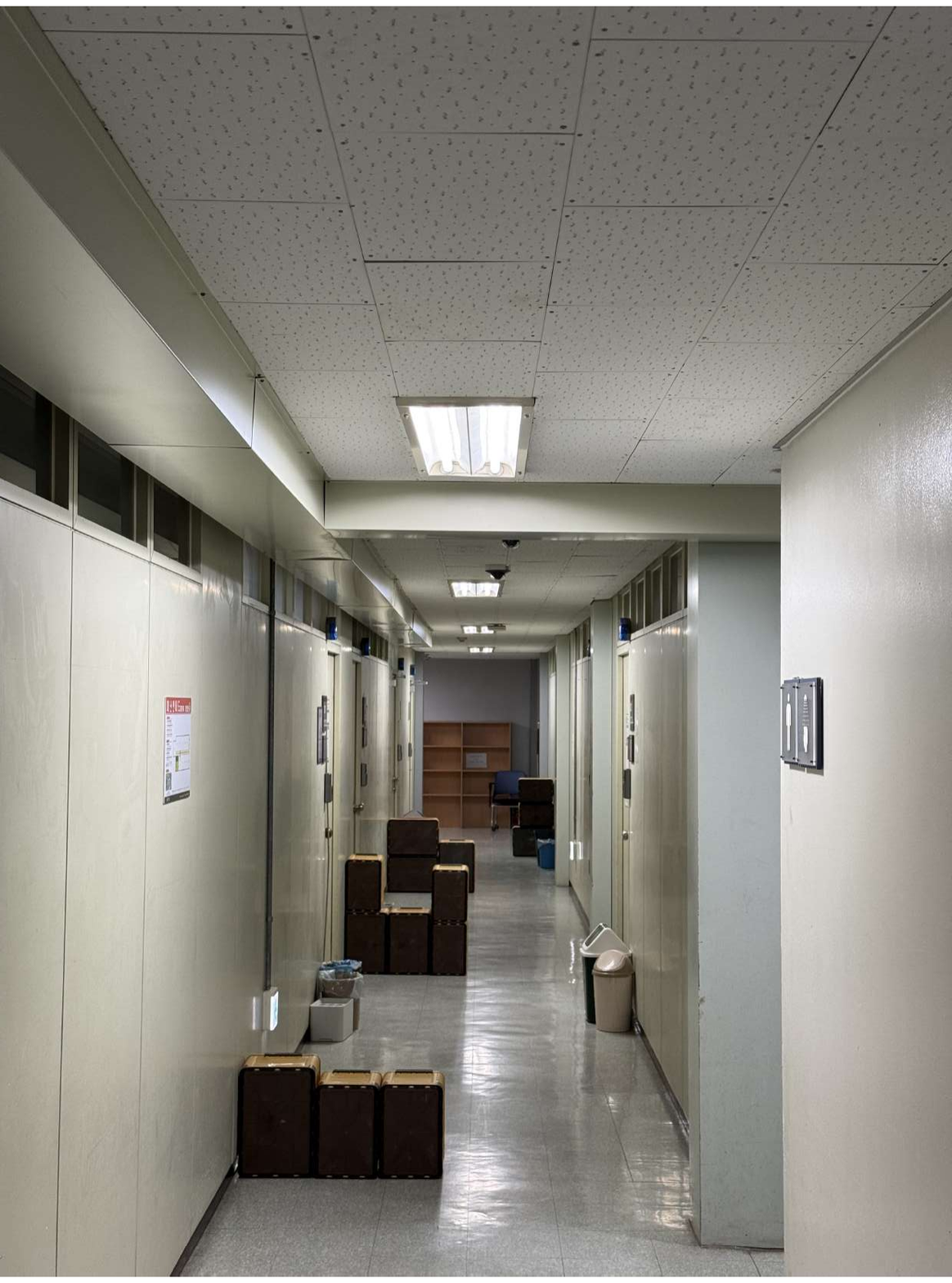}
    \includegraphics[width=0.195\linewidth]{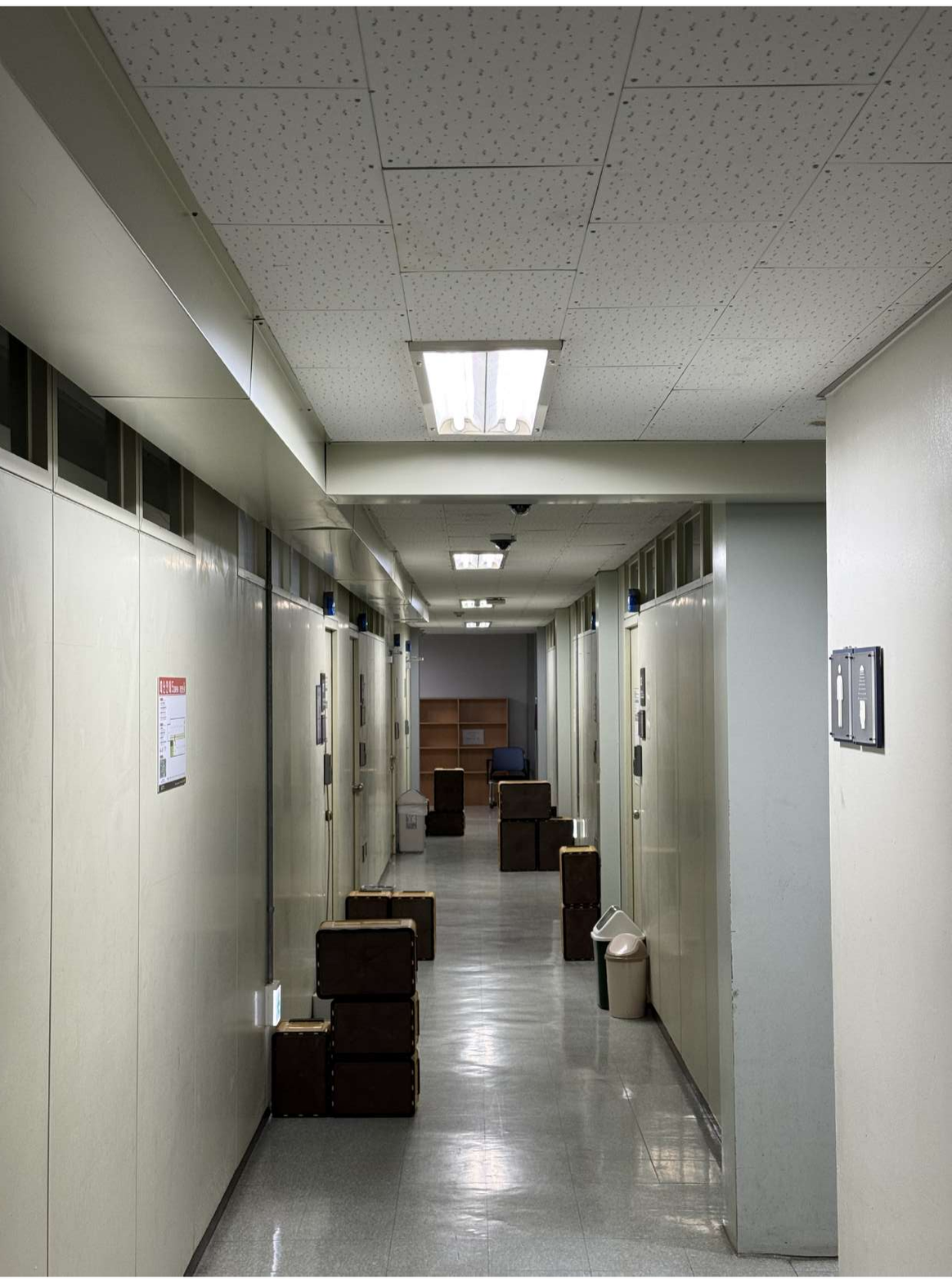}
    \includegraphics[width=0.195\linewidth]{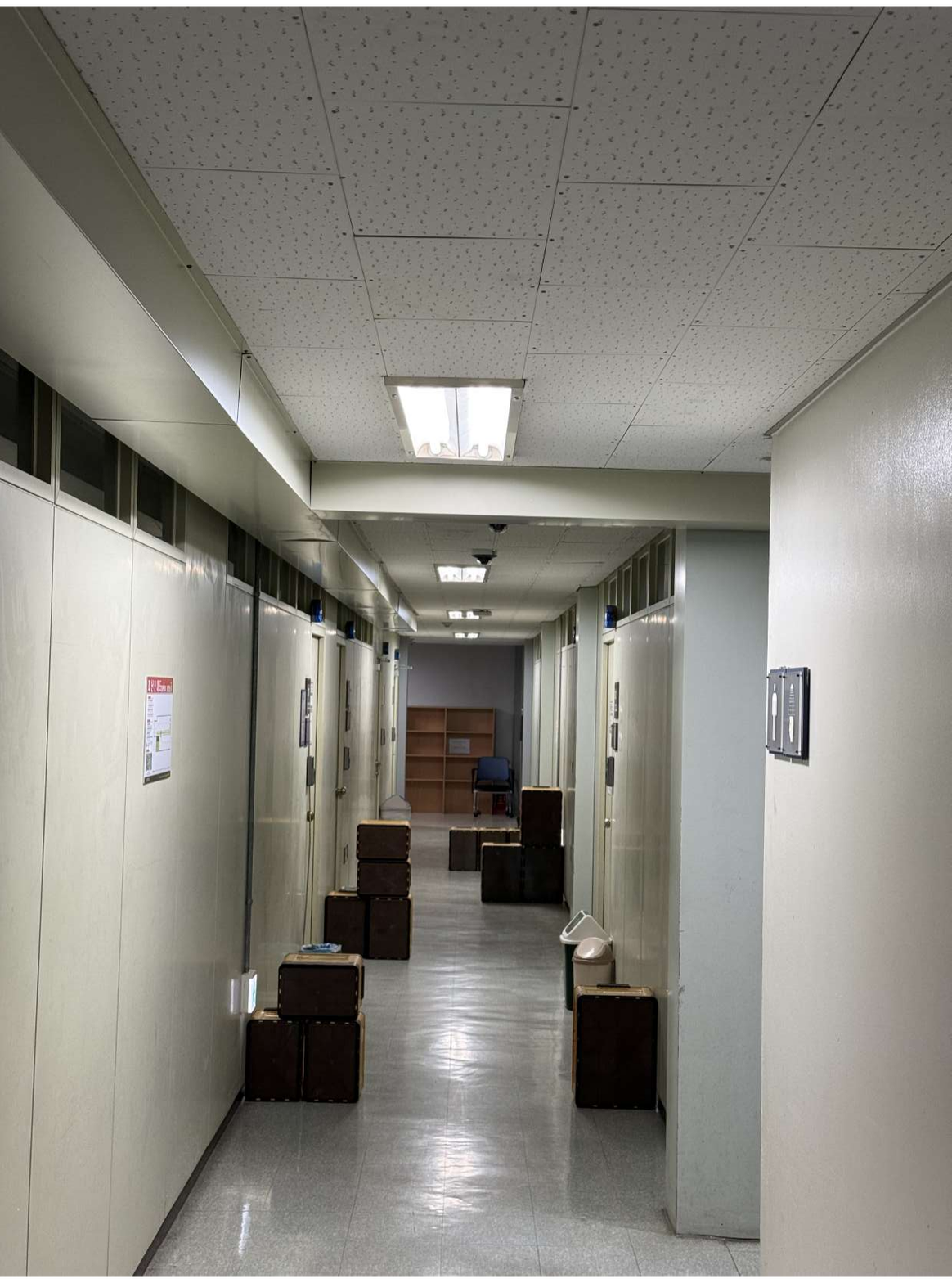}


    \caption{Test instances for goal-conditioned navigation}
    \label{fig:test_instances}
\end{figure*}

Figure~\ref{fig:test_instances} presents the 10 test instances used in the goal-conditioned navigation experiment described in the main paper. Each instance includes 15 OOD obstacles, shown as brown boxes, grouped into four to six clusters of varying shapes and placed randomly throughout the environment. These settings are designed to evaluate the robustness of navigation policies under unseen and cluttered conditions.

\subsection{Robot Platforms and Specifications}
\label{sec:robot_specs}

\begin{figure}[h]
    \centering
    \begin{subfigure}[b]{0.18\linewidth}
        \centering
        \includegraphics[width=\linewidth]{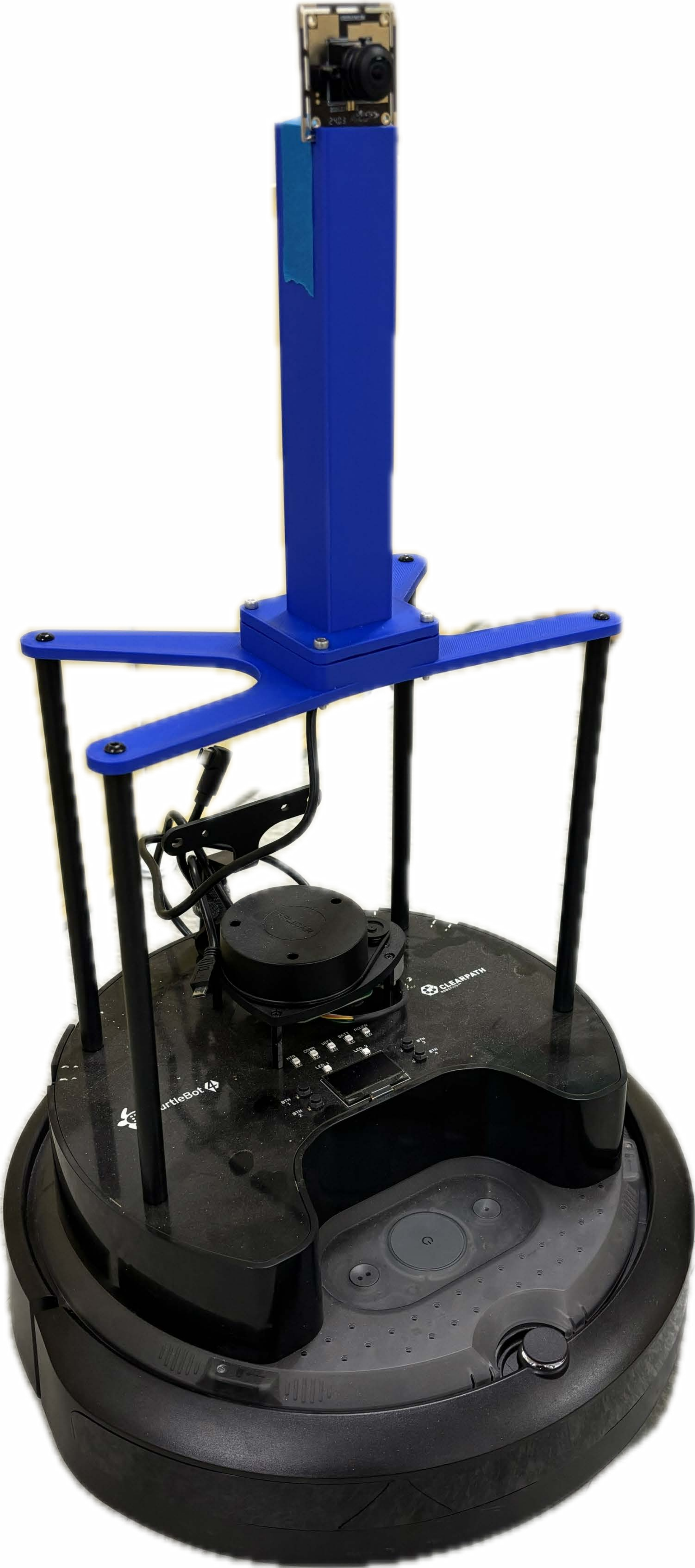}
        \caption{LoCoBot}
    \end{subfigure}
    \hspace{0.75cm}
    \begin{subfigure}[b]{0.18\linewidth}
        \centering
        \includegraphics[width=\linewidth]{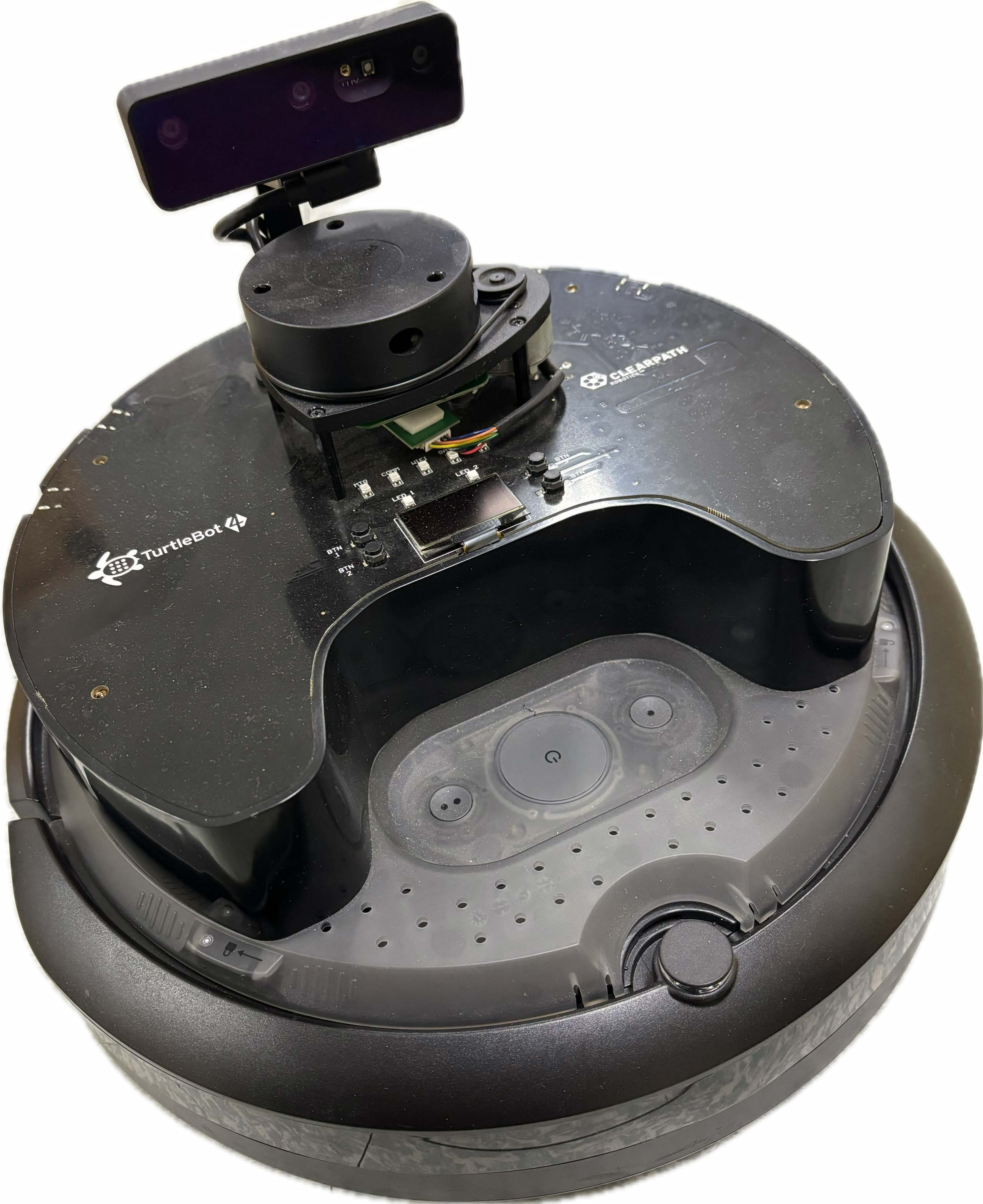}
        \caption{TurtleBot4}
    \end{subfigure}
    \hspace{0.75cm}
    \begin{subfigure}[b]{0.18\linewidth}
        \centering
        \includegraphics[width=\linewidth]{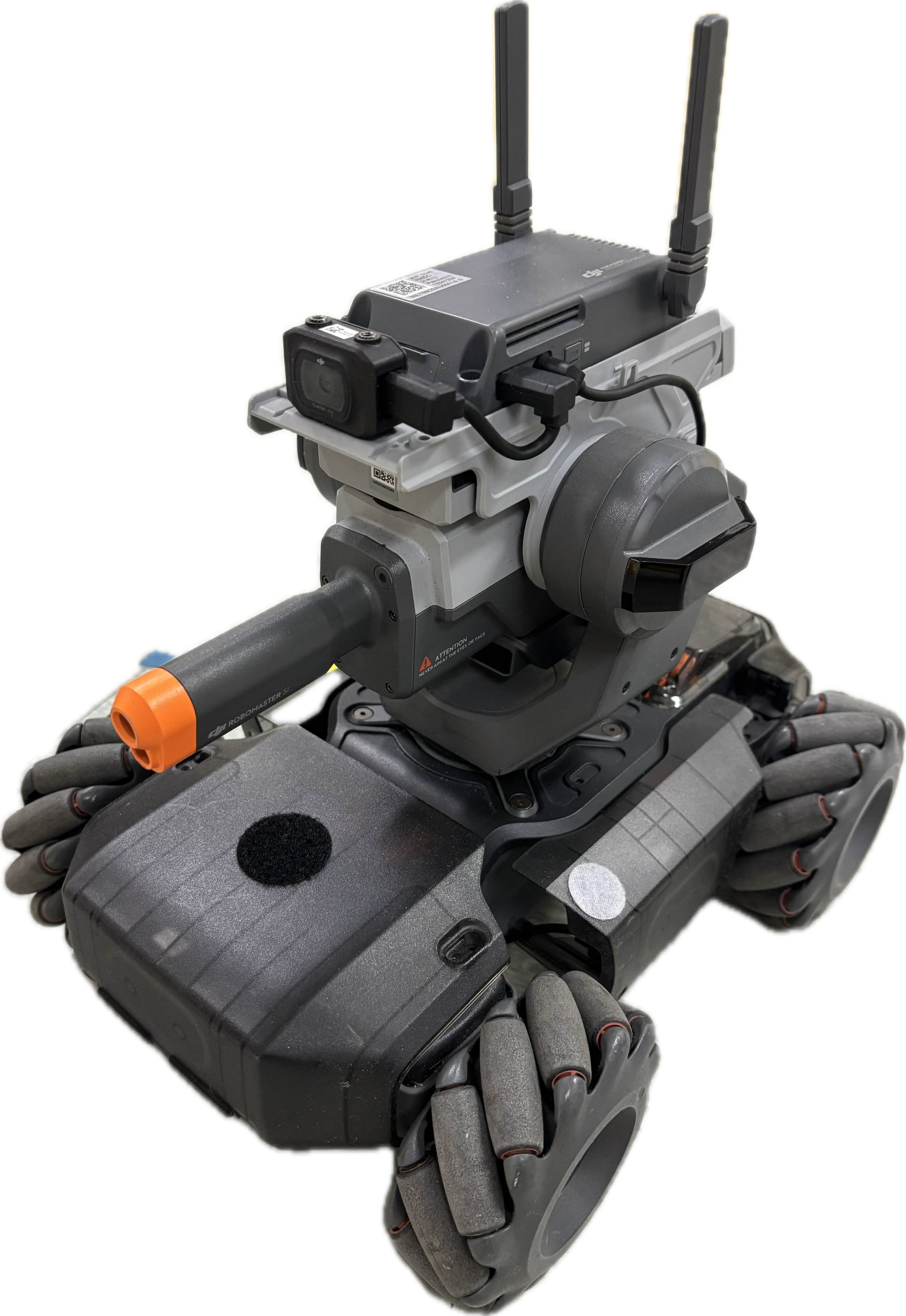}
        \caption{RoboMaster S1}
    \end{subfigure}
    \caption{Mobile robot platforms used for evaluation. Each robot was equipped with a monocular RGB camera.}
    \label{fig:robot_images}
\end{figure}

We run experiments on three mobile platforms: LoCoBot, TurtleBot4, and RoboMaster S1. Figure~\ref{fig:robot_images} shows the actual hardware used in our experiments.

\paragraph{A note for LoCoBot.}
Although the original LoCoBot is built on a TurtleBot2 base, we implement our LoCoBot using a TurtleBot4 as TurtleBot2 has been discontinued. Nevertheless, two models are both differential drive type with very similar wheelbase width, wheel diameter, and etc. A custom 3D-printed camera mount is designed to match the fisheye camera position used in prior NoMaD and ViNT implementations.

\begin{table}[h]
\centering
\caption{Robot specifications and camera configuration. All resolutions are in pixels, and dimensions are in millimeters (mm).}
\label{tab:robot_specs}
\begin{tabular}{@{}lccc@{}}
\toprule
\textbf{Specification} & \textbf{LoCoBot} & \textbf{TurtleBot4} & \textbf{RoboMaster S1} \\
\midrule
Max Depth Range $\tau_z$ (m) & 1.0 & 1.2 & 1.0 \\
Depth Offset (m) & 0.05 & 0.2 & $-0.1$ \\
Published Image Resolution (pixels) & $320 \times 240$ & $320 \times 200$ & $640 \times 360$ \\
Robot Size (L × W × H, mm) & $341 \times 339 \times 350$ & $341 \times 339 \times 351$ & $320 \times 240 \times 270$ \\
Camera Height (mm) & 340 & 245 & 240 \\
Camera X-offset (mm) & 10 & -60 & 70 \\
\bottomrule
\end{tabular}
\end{table}

\paragraph{Notes for parameters in Table~\ref{tab:robot_specs}.}
\begin{itemize}[leftmargin=*]\itemsep-0.01em
    \item Max Depth Range $\tau_z$ (m): The maximum range (in meters) for depth sensing used during obstacle detection. This value is empirically tuned to prevent false positives caused by distant walls or floor misclassification.
    \item Depth Offset (m): An offset subtracted from the estimated depth to make obstacles appear closer (or further) than predicted. This compensates for the physical body radius of robots and camera mounting position, and is further tuned empirically to improve depth estimation.
    \item Published Image Resolution: The resolution of the RGB images published by each camera over ROS 2.
    \item Robot size (L × W × H): Physical dimensions of each robot, measured in millimeters (mm).
    \item Camera Height: The vertical distance from the ground to the optical center of the camera.
    \item Camera X-offset: The horizontal distance (in mm) between the geometric center of robots and the camera. A positive value indicates a forward-facing offset (camera mounted ahead of center), while a negative value indicates a rear-facing offset. 
\end{itemize}


\subsection{Common Parameters for CARE}
\label{sec:care_params}

CARE uses a small set of task-agnostic parameters that are experimentally tuned for effective and safe navigation across all platforms. Table~\ref{tab:care_params} summarizes the key scalar parameters used in the CARE module.

\begin{table}[h!]
\centering
\caption{CARE module parameters used across all experiments.}
\label{tab:care_params}
{\renewcommand{\arraystretch}{1.8}  
\begin{tabular}{@{}llp{7cm}@{}}  
\toprule
\textbf{Parameter} & \textbf{Value} & \textbf{Description} \\
\midrule
$\theta_{\text{clip}}$ (rad) & $\pi/4$ & Maximum heading adjustment angle induced by repulsive force \\
$\theta_{\text{thres}}$ (rad) & $\pi/6$ & Threshold for triggering in-place rotation when the desired heading change exceeds this value \\
$\epsilon$ (m) & $-0.05$ & Vertical offset for filtering out ceiling points during top-down projection. A negative value indicates upward exclusion \\
\bottomrule
\end{tabular}
}
\end{table}

The value of $\theta_{\text{clip}}$ is set to allow sufficient rotation in response to repulsive forces while preventing excessive deviation or backward-pointing waypoints. The value of $\theta_{\text{thres}}$ is chosen to be smaller than $\theta_{\text{clip}}$ to trigger in-place rotation only in high-risk situations with large heading changes. Finally, the vertical offset $\epsilon$ is used to exclude points located above the robot (e.g., ceilings or high shelves) during depth-based top-down projection, improving the relevance of obstacle detection.

\subsection{Depth Estimation Model: UniDepthV2}
\label{sec:depth_model}

For monocular metric depth estimation, we employ UniDepthV2 with the ViT-S backbone (the smallest variant)\footnote{https://github.com/lpiccinelli-eth/UniDepth}, enabling real-time inference. The model is used without any fine-tuning or adaptation. Top-down projections are computed using platform-specific parameters for maximum sensing range $\tau_z$ and vertical offset $\epsilon$, as summarized in Table~\ref{tab:robot_specs} and~\ref{tab:care_params}.

\subsection{Trajectory Selection in NoMaD and ViNT}
\label{sec:traj_selection}

For NoMaD, we use the default diffusion-based sampling setup to generate 8 trajectories, each consisting of 8 waypoints in the robot's local frame. In our experiments, we consistently select the second waypoint $\mathbf{p}_2$ from the first generated trajectory for control (as implemented in the open-source codes). ViNT, on the other hand, produces a single predicted trajectory using a Transformer decoder. Similarly, we select the second waypoint from this predicted trajectory as the control target (also the same with the original implementation of the open-source codes).

This fixed selection strategy is to isolate and evaluate the contribution of our proposed collision avoidance mechanism at the level of local planning. Since our focus is not on high-level trajectory selection or adaptive replanning, we maintain the same trajectory and waypoint index to control for variation across trials. All experiments are conducted under this consistent setting, using fixed random seeds and sampling strategies to ensure fair comparison.

\end{document}